\documentclass[times,referee,twocolumn,final,authoryear]{elsarticle}

\usepackage{framed,multirow}

\usepackage{amssymb}
\usepackage{latexsym}

\usepackage{url}
\usepackage{xcolor}
\definecolor{newcolor}{rgb}{.8,.349,.1}
\newcommand{\resp}[1]{\textcolor{black}{#1}}

\usepackage{graphicx}
\usepackage{subfigure}
\usepackage{booktabs} 
\usepackage{float}
\usepackage{multicol}

\def\ours{{MAEDAY}}
\def\ourss{{\ours{} }}

\begin{document}

\begin{frontmatter}

\title{MAEDAY: MAE for few- and zero-shot AnomalY-Detection}


\author[1,2]{Eli Schwartz\corref{cor1}} 
\cortext[cor1]{Corresponding author}
\ead{eliyahu.schwartz@ibm.com}
\author[1]{Assaf Arbelle}
\author[3]{Leonid Karlinsky}
\author[1]{Sivan Harary}
\author[1]{Florian Scheidegger}
\author[1]{Sivan Doveh}
\author[2]{Raja Giryes}

\address[1]{IBM Research}
\address[2]{Tel-Aviv University, Israel}
\address[3]{MIT-IBM Watson AI Lab}


\begin{abstract}
We propose using Masked Auto-Encoder (MAE), a transformer model self-supervisedly trained on image inpainting, for anomaly detection (AD).
Assuming anomalous regions are harder to reconstruct compared with normal regions.
\ours{} is the first image-reconstruction-based anomaly detection method that utilizes a pre-trained model, enabling its use for Few-Shot Anomaly Detection (FSAD).
We also show the same method works surprisingly well for the novel tasks of Zero-Shot AD (ZSAD) and Zero-Shot Foreign Object Detection (ZSFOD), where no normal samples are available.
\end{abstract}




\end{frontmatter}


\begin{figure*}[h]
    \hspace{-12pt}
    \includegraphics[width=1.05\textwidth]{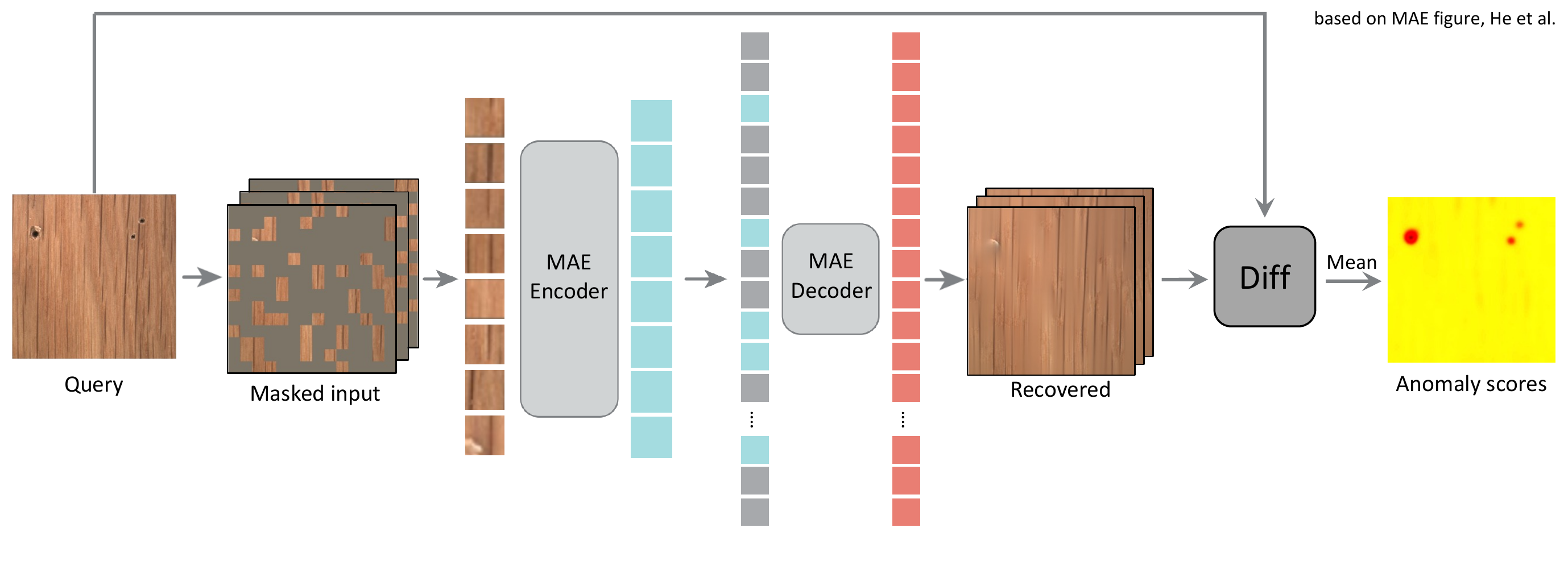}
    \vspace{-25pt}
    \caption{\ours: We repurposed MAE for Zero and Few-Shot Anomaly-Detection. In the zero-shot setup, with no special training and no good images as a reference, ImageNet pre-trained MAE is used to reconstruct a mostly masked-out query image. Anomalous regions are detected in areas where the reconstruction fails, as these regions cannot be accurately inferred from neighboring regions. The anomaly scores are averaged across multiple reconstructions with different random masks. In the few-shot case, the pre-trained model is further finetuned on the reconstruction of the available normal images. Figure adapted from \cite{he2022masked}.}
    \label{fig:teaser}
\end{figure*}

\section{Introduction}
\let\thefootnote\relax\footnotetext{\\Code and data: \url{https://github.com/EliSchwartz/MAEDAY}}
The challenge of Anomaly-Detection (AD) stems from the fact that good cases are similar and easy to model, while anomalies rarely happen, and when they do, they can take an unpredictable form.
For this reason, classic supervised training is sometimes not feasible for AD.
In AD only good images are provided during training,
the goal is to model the distribution of the good images and thus detect outliers at inference time when they occur.
\resp{
Our method is based on image reconstruction, where a model is trained to reconstruct normal images from a corrupted observation, e.g. noisy image or partially masked-out. Assuming that at inference the model will fail to reconstruct anomalous images.}

Recently, there has been a great interest in Few-shot AD (FSAD) \citep{roth2022towards, Sheynin_2021_ICCV, Rudolph_2021_WACV, huang2022registration}.
The promise of FSAD is that a single model can be used for different objects and adapted based on only few good samples.
\resp{
Previous image-reconstruction-based methods are not applicable to the FSAD task since they train the reconstruction model from scratch and therefore require larger training sets.}

We suggest, for the first time, image-reconstruction-based method that can be used for FSAD.
We do that by pre-training the model for general natural-image reconstruction (pre-training on ImageNet).
Our suggested method, \ours, addresses FSAD by using Masked AutoEncoder (MAE) \citep{he2022masked}, a model trained for general image completion based on partial observations, see Fig.~\ref{fig:teaser}.
MAE was introduced for a different purpose, trained on a self-supervised task (image inpainting) with the end goal of learning image representation. 
We re-purpose MAE for FSAD, unlike MAE where the decoder is discarded at inference time, we use both the encoder and the \resp{decoder} to get a recovered image and not just an intermediate representation.
We use the available few good images to further fine-tune the MAE.
The idea is that normal regions will be easier to recover based on patterns observed in the few good examples and based on recurring patterns in the query image itself.
As in the many-shot case, image reconstruction is underperforming compared to embedding-based. 
\resp{However, we observed that an ensemble of SOTA embedding-based method, PatchCore \citep{roth2022towards}, and our reconstruction-based method, \ours, performs extremely well and sets a new state-of-the-art result.}

Following FSAD, we also suggest a new task, Zero-shot AD (ZSAD).
A class-invariant model that takes as input a single query image (without any good reference) and detects anomalies or irregularities.
\resp{Most AD models are trained for a specific object class, some of them are class-invariant in the sense they are trained jointly on all tested classes. In ZSAD the model is pre-trained on unrelated images, not from the classes it is tested on.}
Since the model should detect anomalies with no access to a reference image, it is mostly relevant for textures, where patterns repeat and the query image acts as a self-reference.
Such a model can be particularly useful in industrial settings, e.g. manufacturing of textured materials.
We show that \ours, without any training images, achieves high results for ZSAD and particularly compares favorably to the FSAD SOTA for the textures datasets in MVTec \citep{bergmann2021mvtec}.

We also explore a new task of Zero-Shot Foreign Object Detection (ZSFOD).
Most Foreign Object Detection works are using annotated images with bounding boxes or segmentation masks to train an object-detector \citep{munyer2022foreign,noroozi2023towards,jing2022pixel}.
A common use case is detecting foreign objects or debris on the pavement in airports' runways \citep{munyer2021fod}. 
We focus on the zero-shot case, having a single model that can generalize to new use cases, with no prior reference to either a
free-of-objects surface or the objects to be detected.
We treated this problem similarly to ZSAD where the objects are an
anomaly in the surface texture.
We release a new FOD dataset of wooden floors (indoor) and pavement (outdoor) with or without foreign objects.
We show that \ours, without any training images, outperforms the SOTA one-shot results on this dataset.

To summarise, our contributions are (1) Suggesting \ours{}, MAE-based model pre-trained for image reconstruction on an arbitrary set of images and used for Few-Shot Anomaly-Detection (FSAD); (2) Suggesting the new task of Zero-Shot AD (ZSAD) and demonstrating strong results, particularly for textures (3) Suggesting the new task of Zero-Shot Foreign Object Detection (ZSFOD) and showing strong results; (4) Releasing a new FOD dataset.

\begin{table*}[h!]
\caption{Image-level ROC-AUC results for 0-shot and 1-shot on the MVTec datasets. \ours{} performs surprisingly well even on objects and textures the model was not trained on (ZSAD). In the 1-shot case, the embedding-based method, PC \citep{roth2022towards}, has higher performance when evaluating a single model. However, \ours{} adds a new kind of information and hence a \ours{}+PC ensemble outperforms an ensemble of 2 PC models. Our 1-shot results are presented with mean$\pm$std over 3 different shot selections.}

    \centering
    \begin{tabular}{l|c|cccc|cc}
            \toprule
            & 0-Shot &  \multicolumn{6}{|c}{1-Shot} \\
            \midrule
            & Single-Model  & \multicolumn{4}{|c|}{Single-Model} & \multicolumn{2}{c}{Ensemble} \\
            & \ours & SPADE & PaDiM & PC & \ours & 2*PC & \ours+PC \\
            \midrule
            \multicolumn{6}{l}{\textbf{Objects}} \\
            bottle & 74.3 &&& \textbf{96.1} $\pm$ 3.5 & 74.8  $\pm$ 0.1& \textbf{98.3}  $\pm$ 1.8 &93.7  $\pm$ 1.8 \\
            cable & 53.0 &&& \textbf{82.6} $\pm$ 0.8 & 50.1  $\pm$ 5.0& \textbf{83.6}  $\pm$  2.3&  69.0  $\pm$ 4.6\\
            capsule & 64.0 &&& \textbf{63.0} $\pm$ 1.8& 59.9  $\pm$ 9.5& 63.7 $\pm$ 1.8 &  \textbf{64.9}  $\pm$ 1.9\\
            hazelnut & 97.1 &&& 84.9 $\pm$ 5.6 & \textbf{97.0}  $\pm$ 0.2& 85.4 $\pm$ 5.1 & \textbf{94.1}  $\pm$ 0.2\\
            metal-nut & 43.6 &&& \textbf{75.4} $\pm$ 3.4 & 53.1  $\pm$ 1.5& \textbf{77.0} $\pm$ 2.8 & \textbf{73.4} $\pm$ 1.8 \\
            pill & 63.4 &&& \textbf{77.5} $\pm$ 1.4 & 63.5  $\pm$ 0.5& 79.1  $\pm$ 1.9& \textbf{81.7}  $\pm$ 2.1\\
            screw & 69.9 &&& 46.0 $\pm$ 2.6 & \textbf{78.1}  $\pm$ 2.5& 45.8  $\pm$ 2.6& \textbf{61.4}  $\pm$ 2.2\\
            toothbrush & 77.5 &&& \textbf{84.4} $\pm$ 1.6 & 81.7  $\pm$ 2.9& 83.8 $\pm$ 1.4 & \textbf{92.5}  $\pm$ 1.0\\
            transistor & 48.3 &&& \textbf{82.1} $\pm$ 3.8 & 56.3  $\pm$ 4.1& \textbf{80.1} $\pm$ 5.0 & 75.3   $\pm$ 2.7\\
            zipper & 82.0 &&& \textbf{96.6} $\pm$ 1.4 & 79.0  $\pm$ 0.2& \textbf{96.9} $\pm$ 0.4& 94.3   $\pm$ 1.1\\
            \midrule
            Mean (Objects) & 67.3	&&& \textbf{78.9}	& 69.3 &	79.3 &	\textbf{80.1} \\
            \midrule
            \multicolumn{6}{l}{\textbf{Textures}} \\
            carpet & 74.6 &&& \textbf{99.1} $\pm$ 0.1 & 72.3  $\pm$ 1.1& \textbf{99.2}  $\pm$ 0.0&  97.9  $\pm$ 0.2\\
            grid & 97.9 &&& 43.4 $\pm$ 6.1 & \textbf{97.1}   $\pm$ 0.3& 43.2  $\pm$ 5.5& \textbf{83.9}  $\pm$ 6.5\\
            leather & 92.9 &&& \textbf{100.} $\pm$ 0.0 & 93.4  $\pm$ 0.1& \textbf{100.}  $\pm$ 0.0& 99.9  $\pm$ 0.0\\
            tile & 84.3 &&& \textbf{98.5} $\pm$ 0.2 & 87.2  $\pm$ 1.5& \textbf{98.7}  $\pm$ 0.2& \textbf{98.4}  $\pm$ 0.2\\
            wood & 94.8 &&& \textbf{98.5} $\pm$ 0.5 & 96.7  $\pm$ 0.5 & 98.5  $\pm$ 0.5& \textbf{99.5}  $\pm$ 0.0\\
            \midrule
            Mean (Textures) & 88.9	&&& 87.9	& \textbf{89.3} &	87.9 &	\textbf{95.9} \\
            \midrule
            Mean (All) & 74.5 & 71.6 & 76.1 & \textbf{81.9} & 76.0 & 82.2 & \textbf{85.3} \\
            \bottomrule
    \end{tabular}
    
    \label{tab:mvtec_img}
\end{table*}

\begin{table*}[h!]
\caption{Pixel-level AUC-ROC on MVTec datasets. See Table \ref{tab:mvtec_img} for details.}
    \centering
    \begin{tabular}{l|c|cccc|cc}
            \toprule
            & 0-Shot &  \multicolumn{6}{|c}{1-Shot} \\
            \midrule
            & Single-Model & \multicolumn{4}{|c|}{Single-Model} & \multicolumn{2}{c}{Ensemble} \\
            & MAEDAY & SPADE & PaDiM & PC & \ours & 2*PC & \ours{} + PC \\
            \midrule
            \multicolumn{6}{l}{\textbf{Objects}} \\
bottle & 50.7 &&& \textbf{97.9}  $\pm$ 0.1 & 50.8  $\pm$ 0.5& \textbf{98.1}   $\pm$ 0.1&  95.9  $\pm$ 0.3\\
cable & 65.5 &&& \textbf{90.3}  $\pm$ 1.2& 73.1  $\pm$ 3.1& \textbf{91.3}   $\pm$ 1.0& 84.2 $\pm$ 0.7\\
capsule & 48.1 &&& \textbf{97.1}  $\pm$ 0.1& 48.4  $\pm$ 3.6& \textbf{97.2}   $\pm$ 0.1&  95.3 $\pm$ 1.3\\
hazelnut & 94.1 &&& 88.5  $\pm$ 1.5& \textbf{94.0}  $\pm$ 0.2& 88.8   $\pm$ 1.5& \textbf{98.3} $\pm$ 0.1\\
metal-nut & 39.6 &&& \textbf{89.6}  $\pm$ 0.8& 47.0  $\pm$ 0.7& \textbf{90.1}   $\pm$ 0.6&  68.4 $\pm$ 1.2\\
pill & 61.5 &&& \textbf{94.7}  $\pm$ 0.4& 62.0 $\pm$  1.1& \textbf{95.1}   $\pm$ 0.3&  91.3 $\pm$ 1.2\\
screw & 96.9 &&& 88.6  $\pm$  0.5& \textbf{96.4}  $\pm$ 0.4& 88.8   $\pm$ 0.5&  \textbf{97.4} $\pm$ 0.0\\
toothbrush & 72.3 &&& \textbf{95.0}  $\pm$ 0.2& 77.6  $\pm$ 3.0& \textbf{95.2}   $\pm$ 0.2&  92.2 $\pm$ 0.5\\
transistor & 59.7 &&& \textbf{92.3}  $\pm$ 1.0& 61.9  $\pm$ 0.2& \textbf{92.3}   $\pm$ 0.8&  86.0 $\pm$ 1.9\\
zipper & 76.2 &&& \textbf{96.9}  $\pm$ 0.4& 73.9  $\pm$ 0.6& \textbf{97.1}   $\pm$ 0.3&  96.2 $\pm$ 0.4\\
\midrule
Mean (Objects) & 66.5 &&& 93.0 & 69.9 & 93.3 & 90.5\\
\midrule
\multicolumn{6}{l}{\textbf{Textures}} \\
carpet & 76.2 &&& \textbf{98.9}  $\pm$ 0.0& 78.4  $\pm$ 1.7& \textbf{99.0}   $\pm$ 0.0& 98.2 $\pm$ 0.2\\
grid & 95.4 &&& 55.7  $\pm$ 0.3& \textbf{96.7}  $\pm$ 0.3& 55.9  $\pm$ 0.3 & \textbf{96.6} $\pm$ 0.2\\
leather & 94.6 &&& \textbf{99.1} $\pm$ 0.0 & 96.4  $\pm$ 0.5& 99.1   $\pm$ 0.0& \textbf{99.4} $\pm$ 0.0\\
tile & 30.9 &&& \textbf{94.8} $\pm$ 0.5 & 37.4  $\pm$ 2.1& \textbf{94.9}   $\pm$ 0.5&  90.1 $\pm$ 1.1\\
wood & 78.8 &&& \textbf{92.0}  $\pm$ 0.2& 80.0  $\pm$ 0.4& 92.1   $\pm$ 0.2& \textbf{92.9} $\pm$ 0.4\\
\midrule
Mean (Textures) & 75.2 &&& 88.1 & 79.7 & 88.2 & 95.4\\ 
\midrule
Mean & 69.4 & \textbf{91.9} & 88.2 & 91.4 & 71.6 & 91.7 & \textbf{92.2}\\
            \bottomrule
    \end{tabular}
    
    \label{tab:mvtec_pixel}
\end{table*}

\section{Related Work}

AD methods divide into two categories: embedding-similarity-based and image-reconstruction-based.

Embedding-similarity-based methods compare image or patch embedding with a distribution of normal image or patch embeddings (modeled by the training set), e.g. \citep{roth2022towards,defard2021padim,cohen2020sub, huang2022registration}.
Some methods perform registration of the images, i.e. spatial mapping of the image to some canonical form \citep{huang2022registration,chen1999anomaly}.
Other approaches learn the negative distribution, too. That requires some assumptions on the anomaly distribution and is achieved by artificially producing anomalies \citep{zou2022spot,li2021cutpaste}.
Following the great interest in Few-Shot object-recognition \citep{vinyals2016matching,snell2017prototypical,doveh2021metadapt}, recently Few-Shot AD (FSAD) also gained popularity.
Similarity-based methods demonstrated success in this low data regime (FSDA) thanks to their use of pre-trained models \citep{roth2022towards, Sheynin_2021_ICCV, Rudolph_2021_WACV}. 
We suggest using pre-trained models for image-reconstruction-based methods as well.

Image-reconstruction-based methods usually train a generative model on a set of normal images, e.g. an AutoEncoder \citep{hinton1990connectionist,japkowicz1995novelty,sakurada2014anomaly} or GAN \citep{goodfellow2014generative,schlegl2017unsupervised,zenati2018efficient,xia2022gan}. The \resp{underlying assumption} is that only good images can be generated by the trained model.
Another kind of generative-model is Normalizing Flows \citep{rezende2015variational}, by using an invertible mapping from a latent space with controlled distribution to images we also obtain the inverse mapping that allows verifying the likelihood of a query image \citep{yu2021fastflow,gudovskiy2022cflow,zhang2021self}.
Other methods apply some form of image degradation and again train a model to reconstruct the images, assuming only good images will be well-reconstructed \citep{yan2021learning,fei2020attribute,zavrtanik2021reconstruction,wyatt2022anoddpm}. 
\resp{The closest approach to ours is RIAD \citep{zavrtanik2021reconstruction}, which masks parts of the image and performs image inpainting.} 
However, RIAD and other image-reconstruction methods rely on training a model from scratch on normal images and are not intended for the low-data or no-data regime.

\resp{Finally, concurrently with our work, new works have started to explore zero-shot anomaly detection. These works are based on pretrained vision-language models and utilize textual prompts to detect anomalies in a class-invariant manner \citep{jeong2023winclip,cao2023segment}.}


\section{Method}
We begin by describing our approach (\ours) for ZSAD which is based on image reconstruction from partial observations. MAE~\citep{he2022masked} is trained on the self-supervised task of predicting an image from a partial observation. This makes MAE a great tool for our purpose. We use an ImageNet pretrained MAE as our backbone.

As commonly done in transformer-based architectures, the input image $I$ is split into non-overlapping patches, and each patch is flattened into a single token. The tokens go through a linear projection with the addition of a positional encoding and are then processed by a sequence of transformer blocks. For MAE most of the input tokens are masked out and discarded, therefore the encoder operates on a small number of tokens. The decoder receives the output tokens of the encoder and in addition `empty' tokens with just the positional encoding replacing the masked-out tokens. Through a sequence of transformer blocks, the decoder `fills' these empty tokens based on information from the encoder output tokens. The output of the decoder is the recovered image.

Usually, at inference time only the MAE encoder is used (for features extraction), while the decoder is discarded. In our case, we use both the encoder and decoder. Given a query image, a random small subset of its patches ($25\%$) are fed to the MAE. The recovered image is then compared against the query image and mismatched pixels indicate an anomalous region. We repeat this process multiple times for each image, each time a different subset of the tokens is retained. With enough repetitions (we used $N=32$) each token is likely to be masked out at least once, such that we can measure how well it is reconstructed. 
\resp{Multiple reconstruction attempts (with different random masks) has also an advantage in cases of ambiguity, with multiple plausible image completions.
In those cases, the model will likely choose the correct completion at least in some of the reconstruction attempts.}
We found in our experiments that the reconstruction for retained tokens (not masked-out) is also somewhat indicative of them being normal vs. anomalous. Our intuition for that is that since the transformer mixes the information from all tokens, even when a token is visible it will be better reconstructed when it is in agreement with its surrounding tokens. Given this observation, we can simply run a query image $N$ times with different random masks and compare the $N$ reconstructed images (full images) against the query image. The method is illustrated in Figure \ref{fig:teaser}.

Formally, given a query image $I \in \mathbb{R}^{H \times W \times3}$ and a set of $N$ random masks $\{M_1,...,M_N\}$, we use MAE to get $N$ reconstructed images $\{R_1,...,R_N\}$, where $R_i=MAE(I\cdot M_i)$. Image resolution and patch size are the same as those used for pretraining MAE (224 and 16).
\resp{Then, $\{R_i\}$ are used to compute $N$ squared error maps.}
The squared error maps are channel-wise filtered with a Gaussian kernel $g$ (kernel size 7, $\sigma=1.4$) to remove noise and summed over the 3 color channels, 
\begin{equation}
E_i=\sum_{c\in\{R,G,B\}}{(I^c-R_i^c)^2 \ast g}.
\end{equation}

The $N$ error maps are averaged to get a single error map, $E=\frac{1}{N} \sum_{i=1}^{N}{E_i}$.
$E$ is the pixel-level anomaly score.
Finally, the image anomaly score is set by the max error $S=max(E)$.

For FSAD we first finetune the MAE model with the available normal images.
Unlike MAE, where the loss is applied only on the recovered masked-out patches, we apply the loss to all patches.
We do that because we use all predicted patches (both masked and unmasked) for detecting anomalies.
We use LoRA \citep{hu2021lora}, a method originally introduced for finetuning large language models (transformers) without overfitting a small dataset. In LoRA additional low-rank weight matrix is introduced for each weight matrix in the original pre-trained model. The low rank is enforced by having a low-rank decomposition. During fine-tuning, only the low-rank weights are updated and the output of each multiplication is the sum of performing the multiplication with the original weights and the new low-rank weights. After finetuning is finished, the weights are updated to be the sum of the original weights and the new ones (to avoid additional compute and memory consumption at inference time). 

We set the rank of the additional LoRA weights to $32$ for all tensors in the model.
The model is trained for 50 iterations using an SGD optimizer with a learning rate of $1e-2$ (LoRA requires a relatively high learning rate), a momentum of $0.9$, and a weight decay of $0.05$. We train with random crop and random rotation augmentations. The batch size is set to 32, so the few available shots are used multiple times to fill the batch (but with different random masks each time).

\section{Results}

\begin{figure*}[htp]
    \centering
    \begin{tabular}{cc}
         \includegraphics[width=0.3\linewidth]{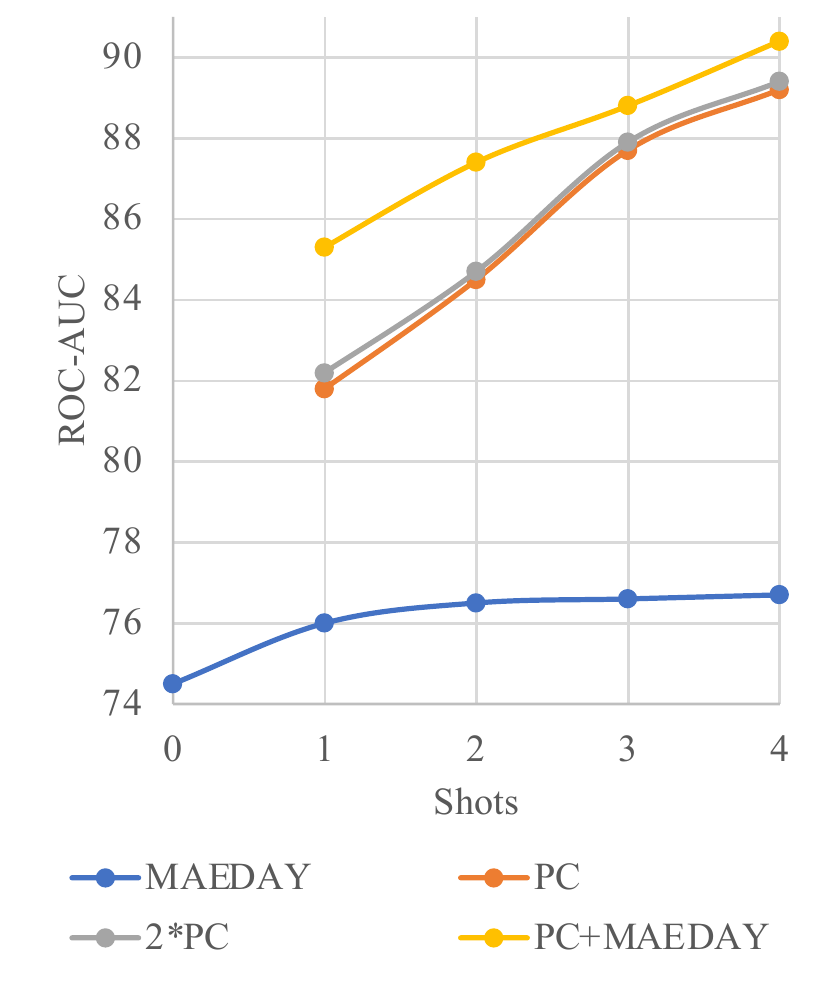} & 
         \includegraphics[width=0.3\linewidth]{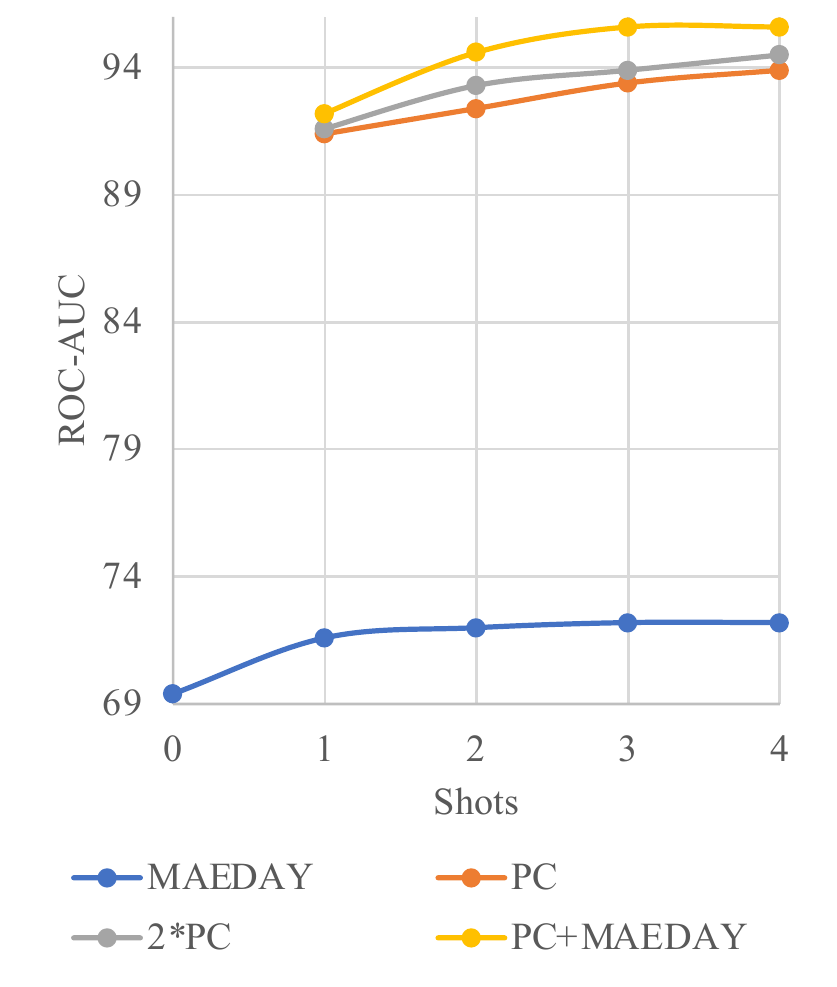} \\
         Image-level ROC-AUC & Pixel-level ROC-AUC
    \end{tabular}
    
    \caption{ROC-AUC for 0-4 shot on the MVTec dataset.}
    \label{fig:more_shots}
\end{figure*}



\begin{figure}[htp]
    \centering
    \includegraphics[width=0.8\linewidth]{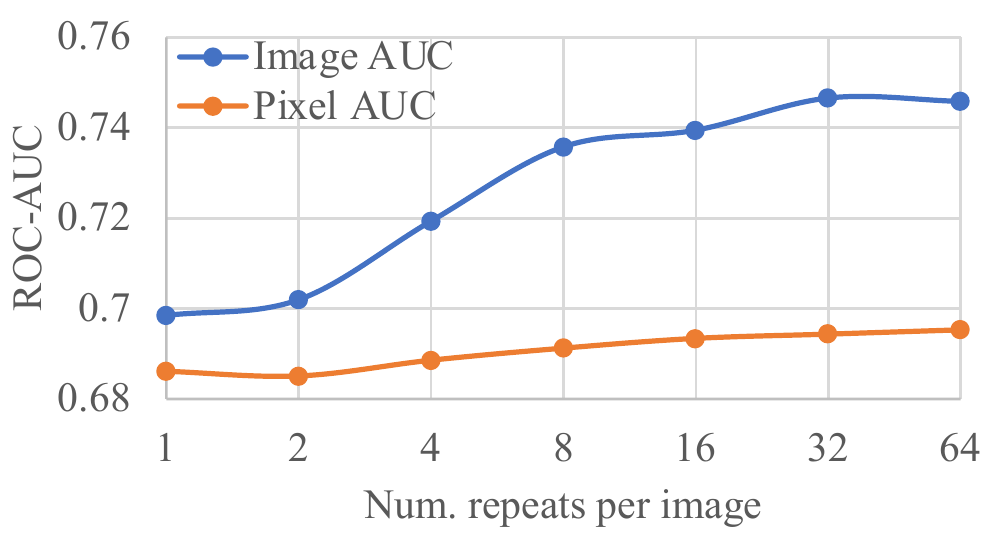}
    \caption{\textbf{Number of repetitions per image}. Scores for each image are averaged over multiple reconstructions with different random masks. We observe performance saturation at $\sim32$ repetitions.}
    \label{fig:num_reps}
\end{figure}

\begin{figure}[h!]
    \centering
    \resizebox{0.96\linewidth}{!}{
    \begin{tabular}{cc|ccc}
        \multicolumn{2}{c|}{Normal} & \multicolumn{3}{c}{Anomaly} \\
        Query & Predicted & Query & Predicted & GT \\
\includegraphics[width=0.2\linewidth]{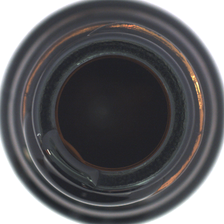} & 
\includegraphics[width=0.2\linewidth]{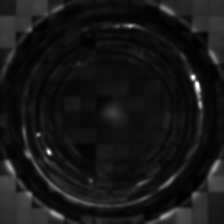} &
\includegraphics[width=0.2\linewidth]{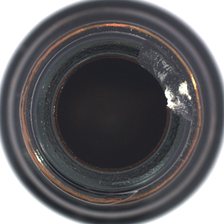} & 
\includegraphics[width=0.2\linewidth]{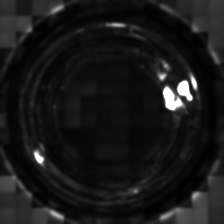} & 
\includegraphics[width=0.2\linewidth]{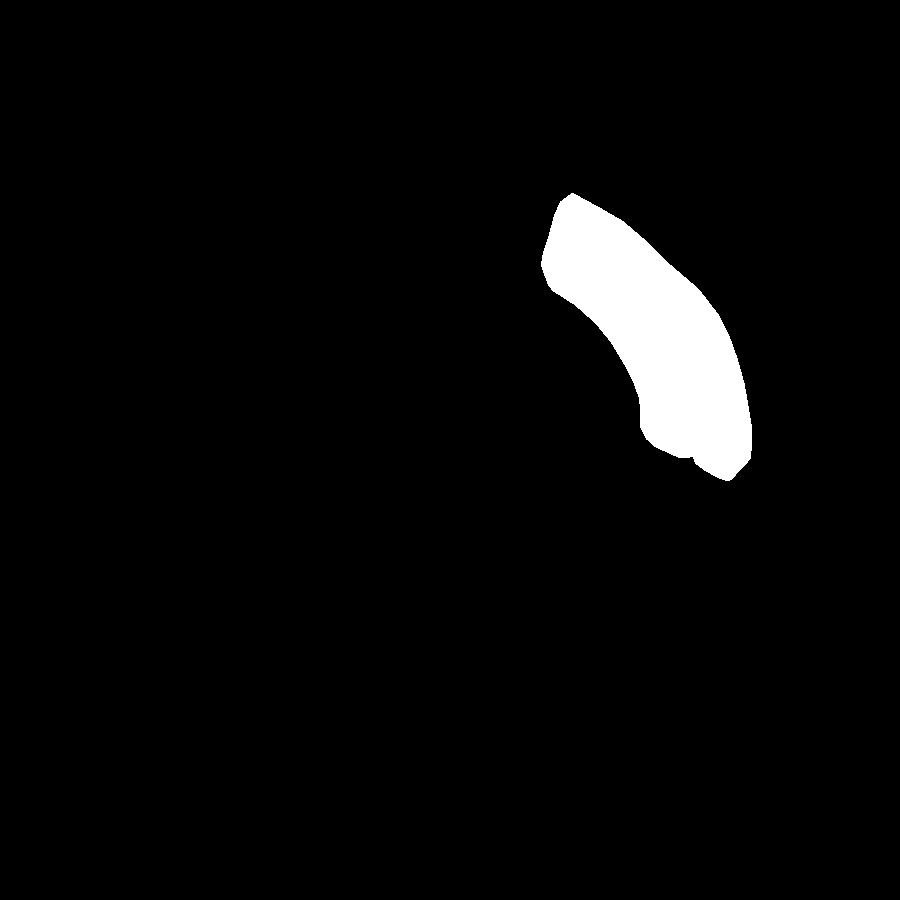}
\\
\includegraphics[width=0.2\linewidth]{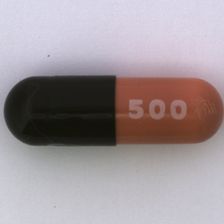} & 
\includegraphics[width=0.2\linewidth]{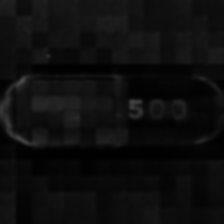} &
\includegraphics[width=0.2\linewidth]{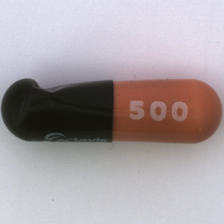} & 
\includegraphics[width=0.2\linewidth]{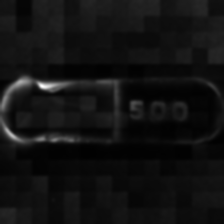}  & 
\includegraphics[width=0.2\linewidth]{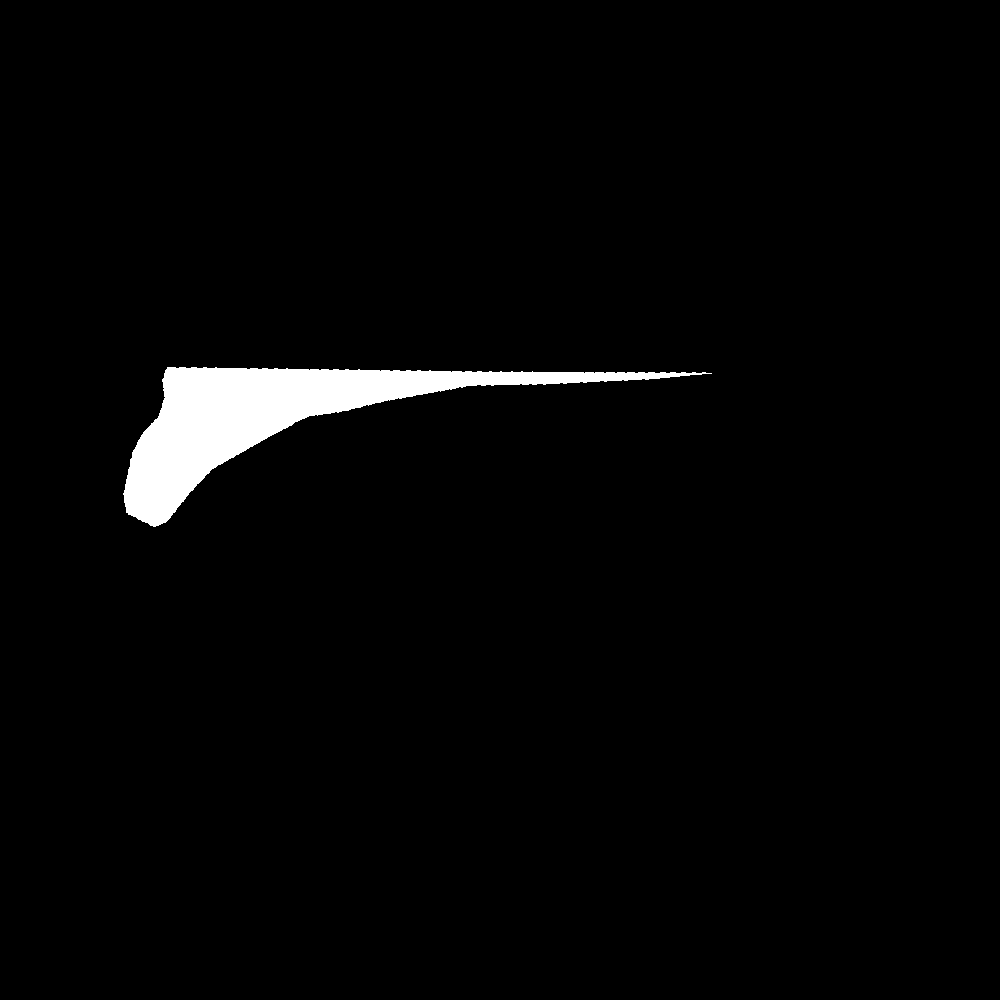} 
\\
\includegraphics[width=0.2\linewidth]{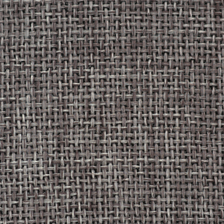} & 
\includegraphics[width=0.2\linewidth]{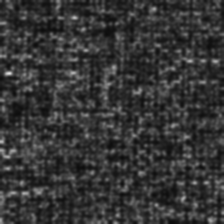} &
\includegraphics[width=0.2\linewidth]{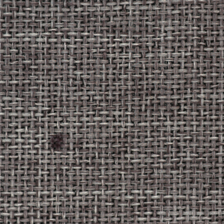} & 
\includegraphics[width=0.2\linewidth]{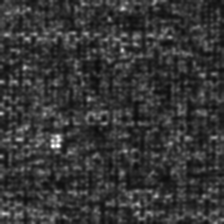}  & 
\includegraphics[width=0.2\linewidth]{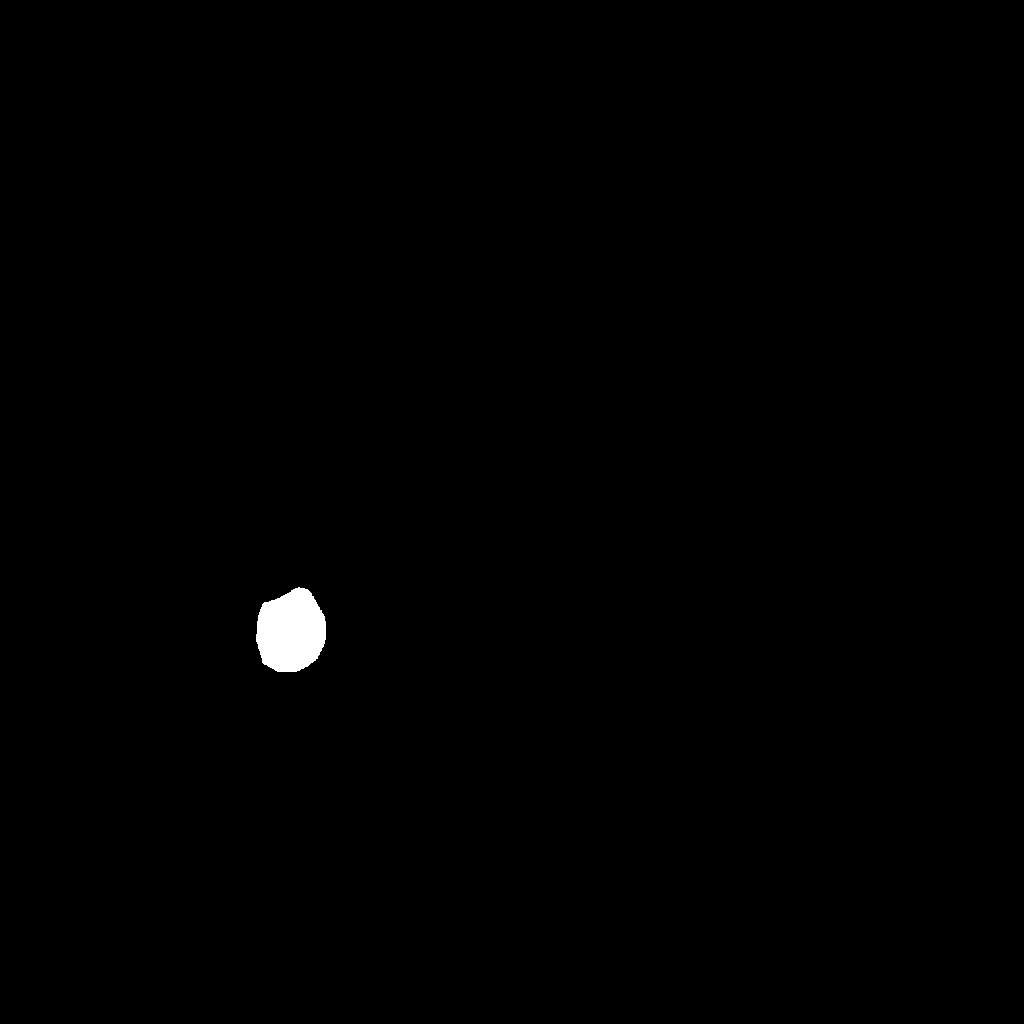} 
\\
\includegraphics[width=0.2\linewidth]{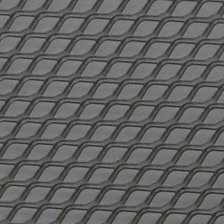} & 
\includegraphics[width=0.2\linewidth]{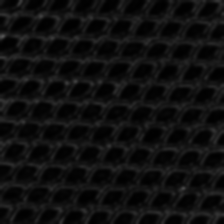} &
\includegraphics[width=0.2\linewidth]{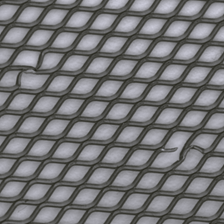} & 
\includegraphics[width=0.2\linewidth]{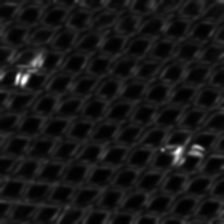}  & 
\includegraphics[width=0.2\linewidth]{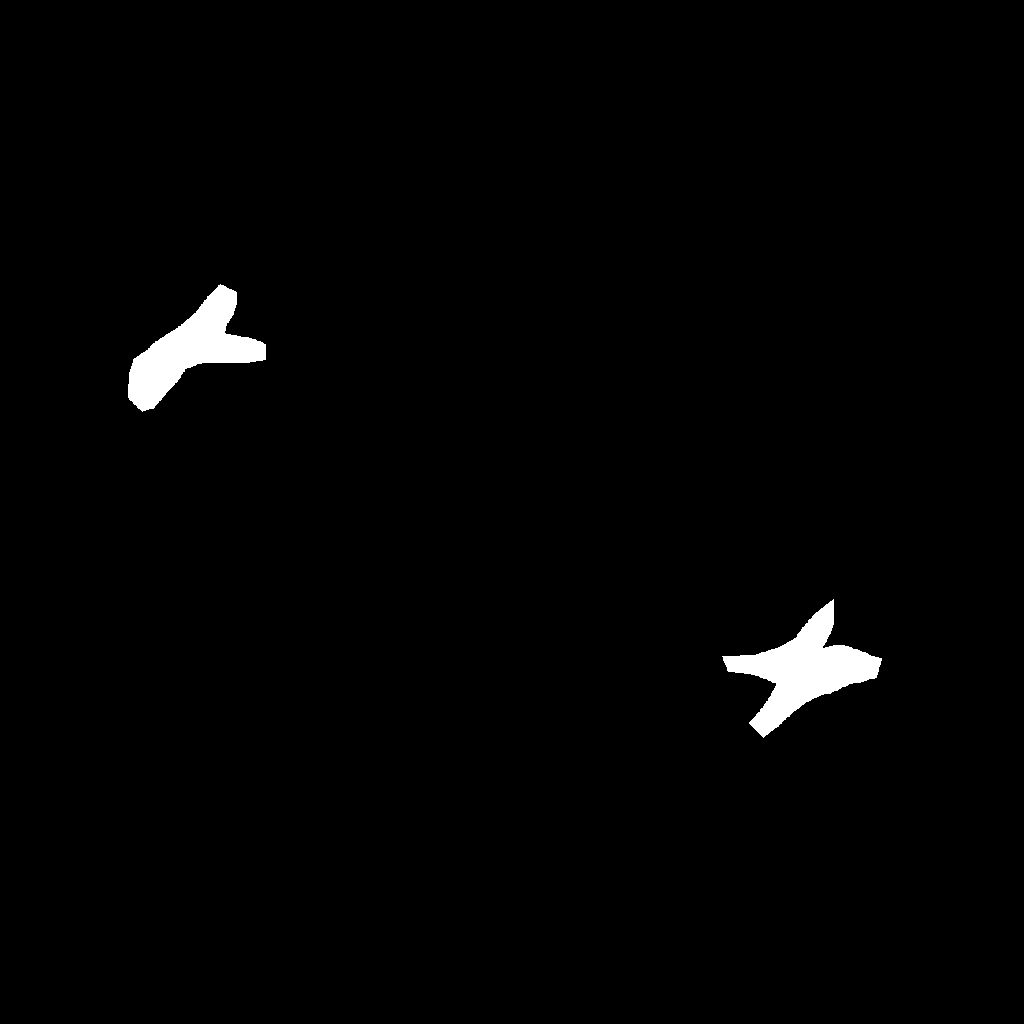} 
\\
\includegraphics[width=0.2\linewidth]{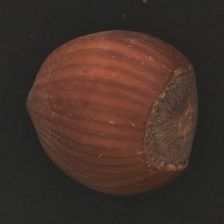} & 
\includegraphics[width=0.2\linewidth]{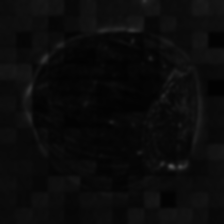} &
\includegraphics[width=0.2\linewidth]{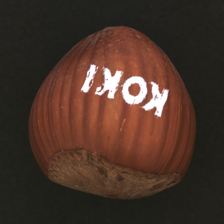} & 
\includegraphics[width=0.2\linewidth]{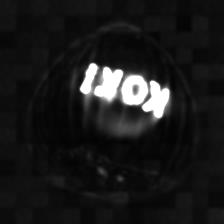}  & 
\includegraphics[width=0.2\linewidth]{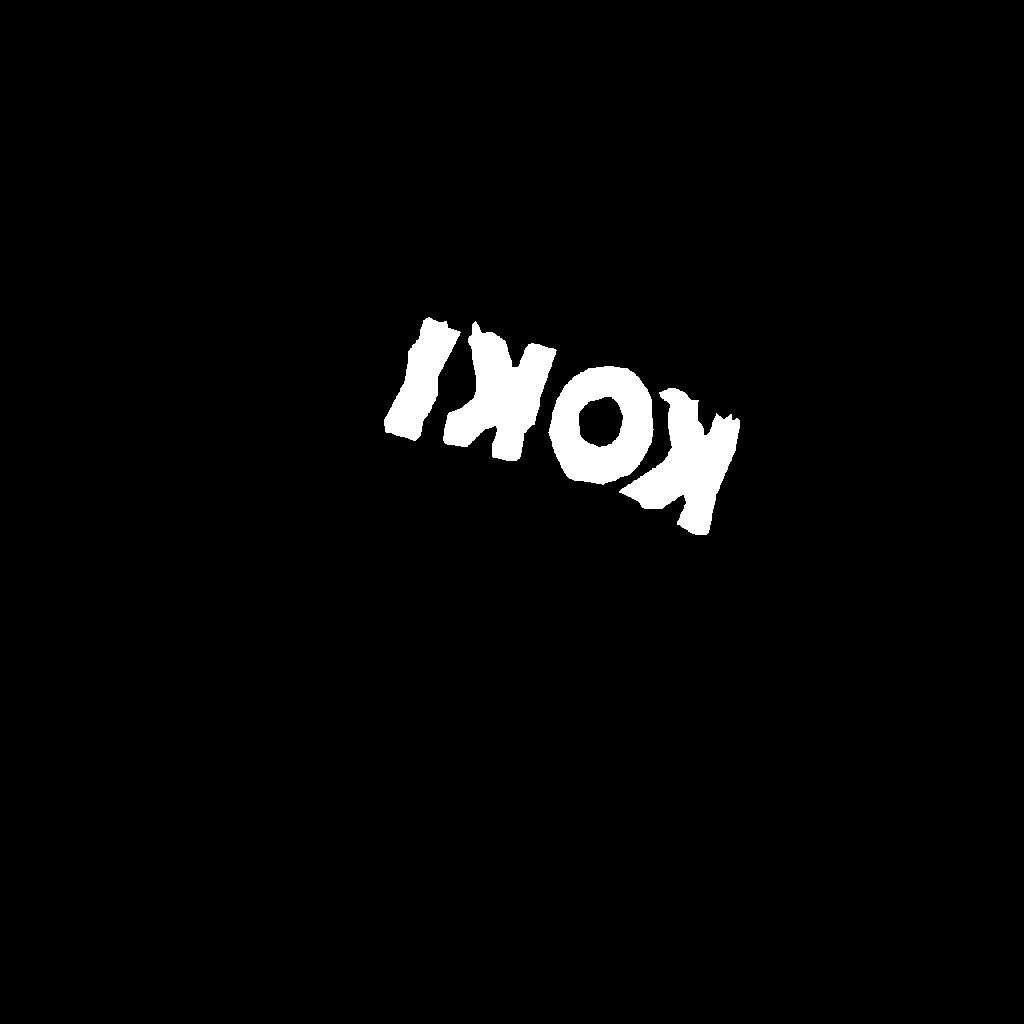} 
\\
\includegraphics[width=0.2\linewidth]{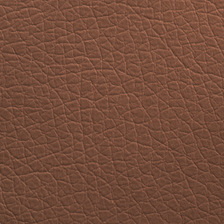} & 
\includegraphics[width=0.2\linewidth]{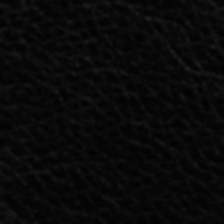} &
\includegraphics[width=0.2\linewidth]{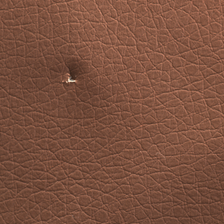} & 
\includegraphics[width=0.2\linewidth]{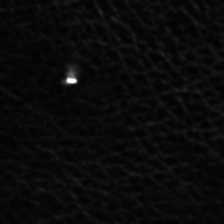}  & 
\includegraphics[width=0.2\linewidth]{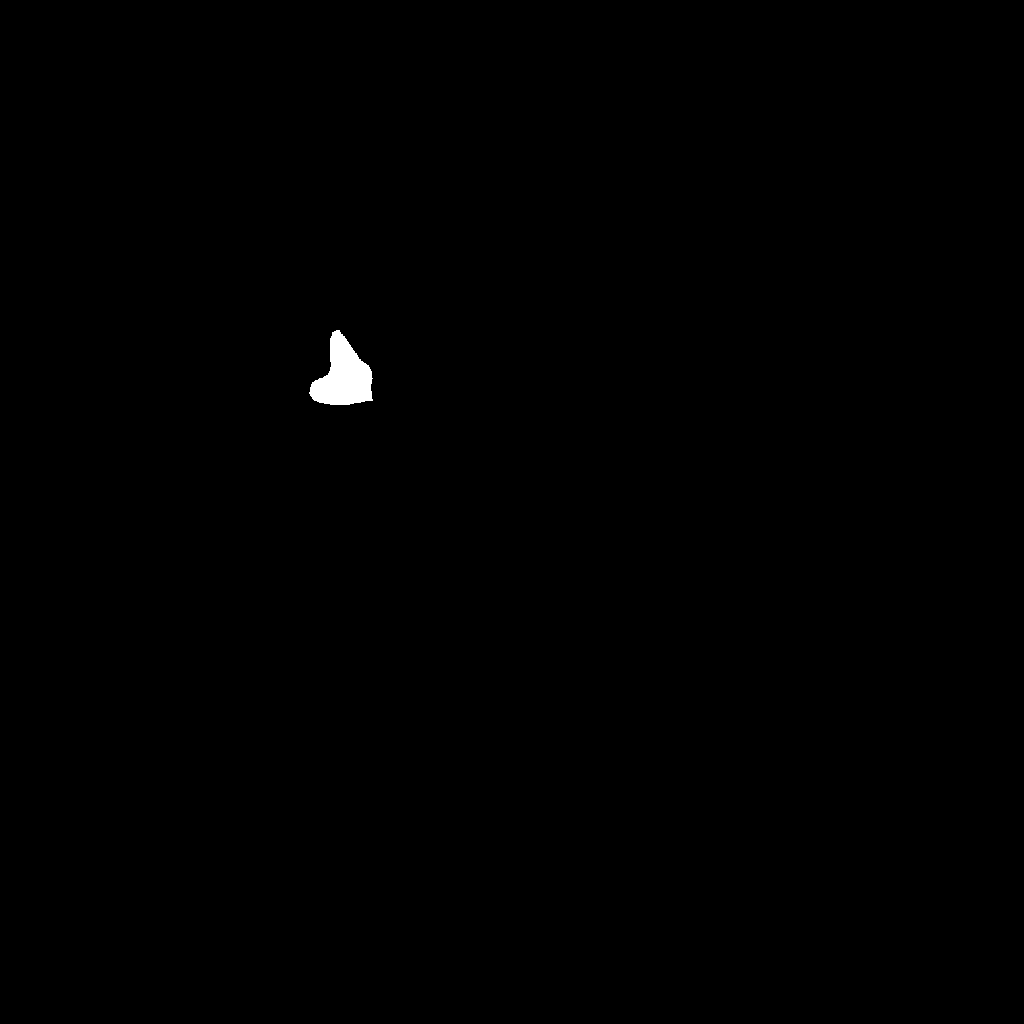} 
\\
\includegraphics[width=0.2\linewidth]{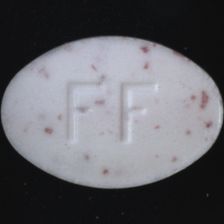} & 
\includegraphics[width=0.2\linewidth]{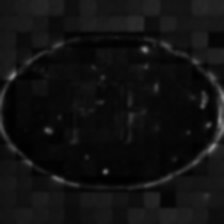} &
\includegraphics[width=0.2\linewidth]{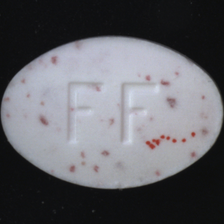} & 
\includegraphics[width=0.2\linewidth]{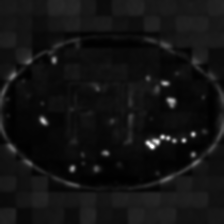}  & 
\includegraphics[width=0.2\linewidth]{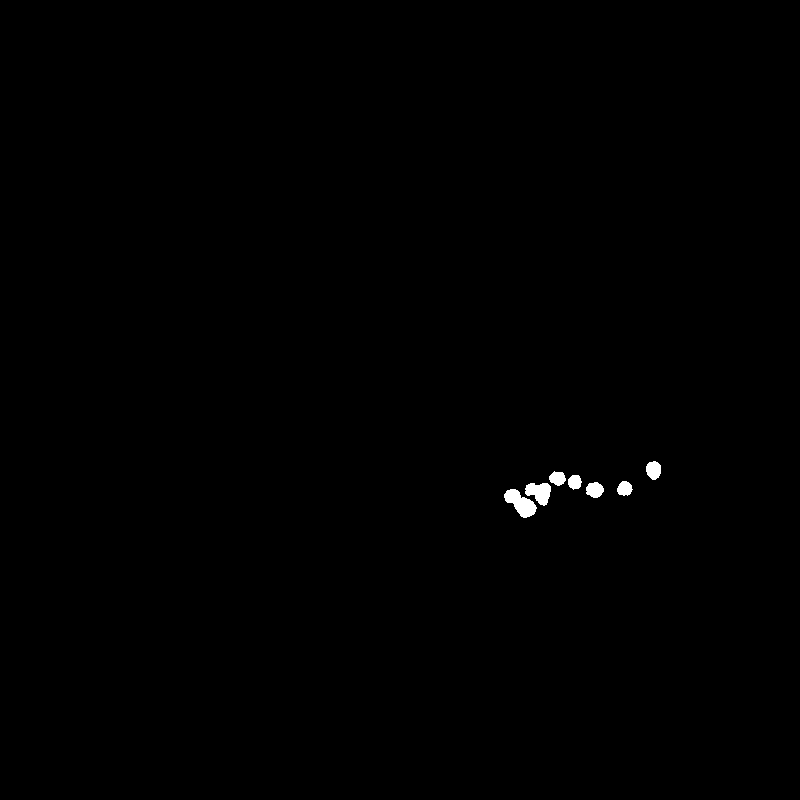} 
\\
\includegraphics[width=0.2\linewidth]{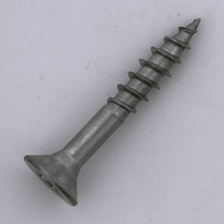} & 
\includegraphics[width=0.2\linewidth]{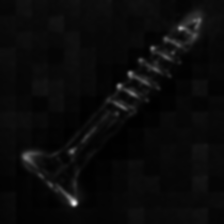} &
\includegraphics[width=0.2\linewidth]{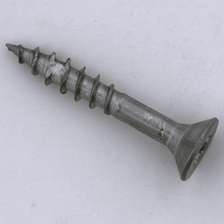} & 
\includegraphics[width=0.2\linewidth]{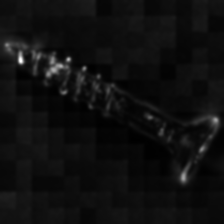}  & 
\includegraphics[width=0.2\linewidth]{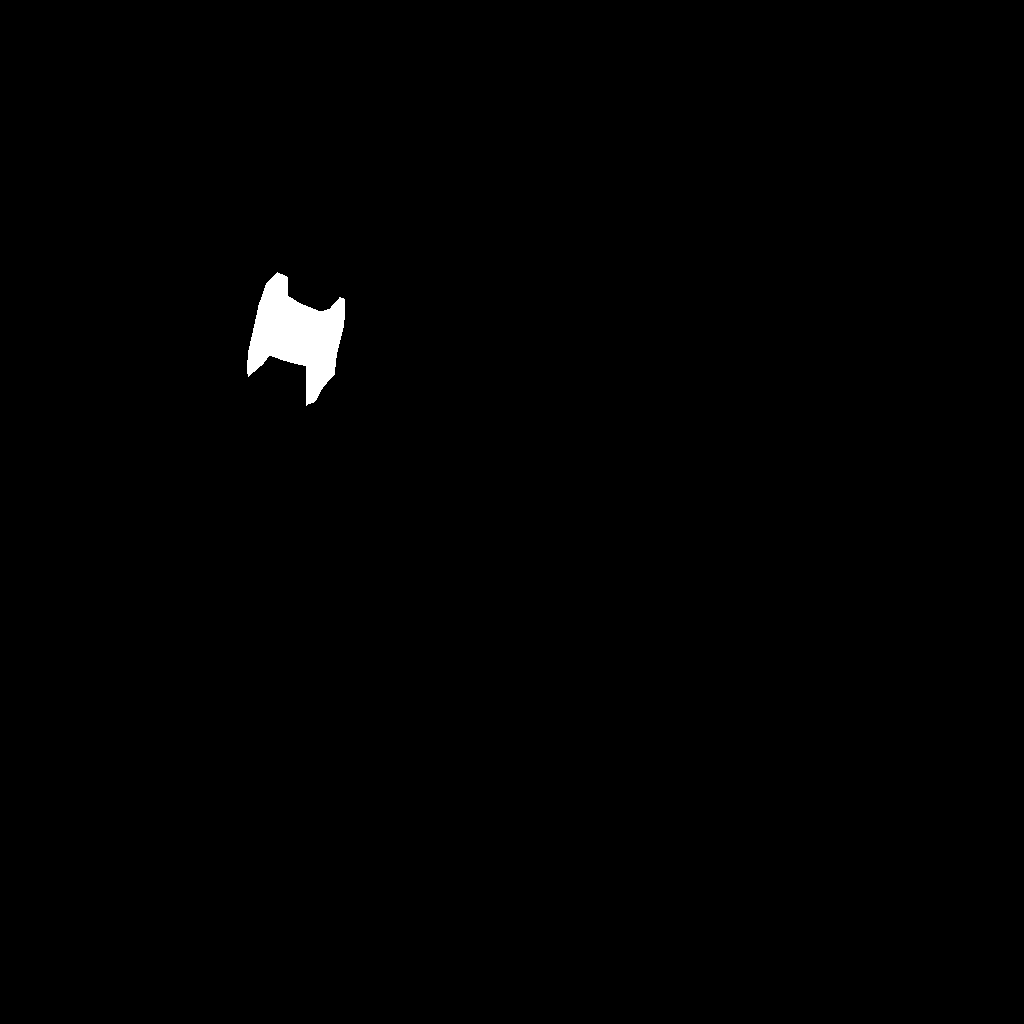} 
\\
\includegraphics[width=0.2\linewidth]{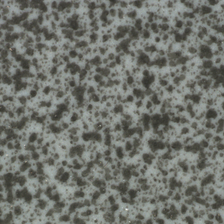} & 
\includegraphics[width=0.2\linewidth]{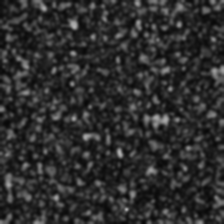} &
\includegraphics[width=0.2\linewidth]{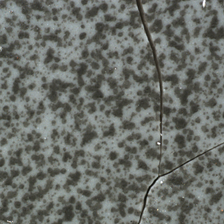} & 
\includegraphics[width=0.2\linewidth]{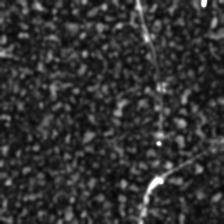}  & 
\includegraphics[width=0.2\linewidth]{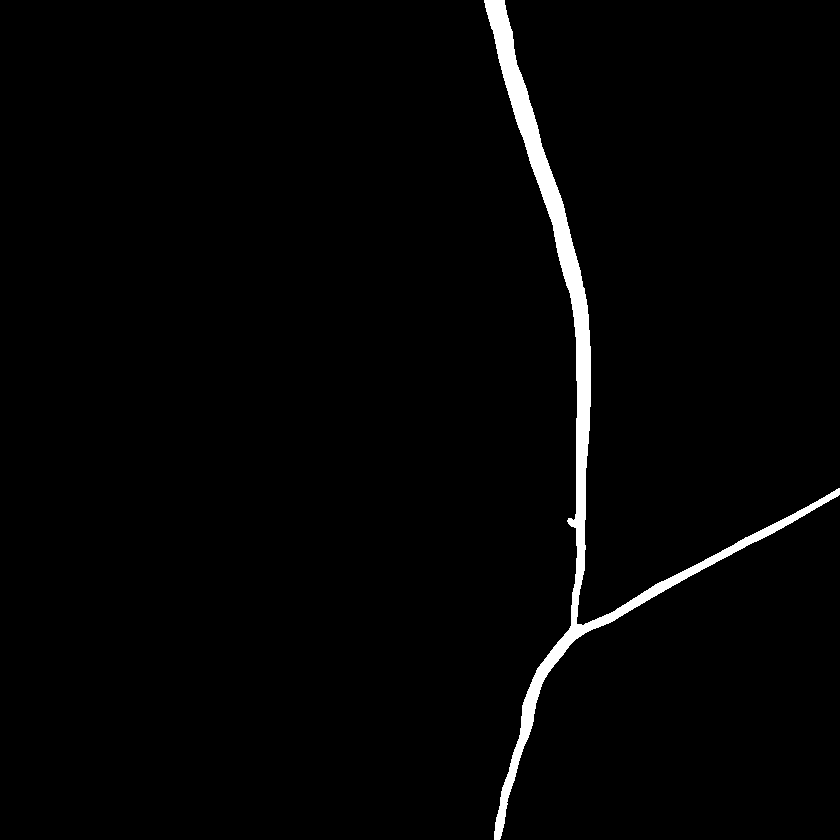} 
\\
\includegraphics[width=0.2\linewidth]{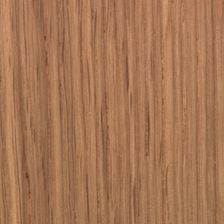} & 
\includegraphics[width=0.2\linewidth]{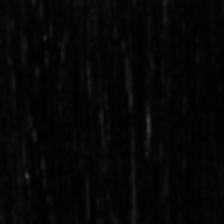} &
\includegraphics[width=0.2\linewidth]{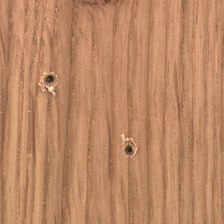} & 
\includegraphics[width=0.2\linewidth]{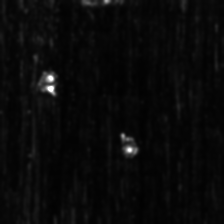}  & 
\includegraphics[width=0.2\linewidth]{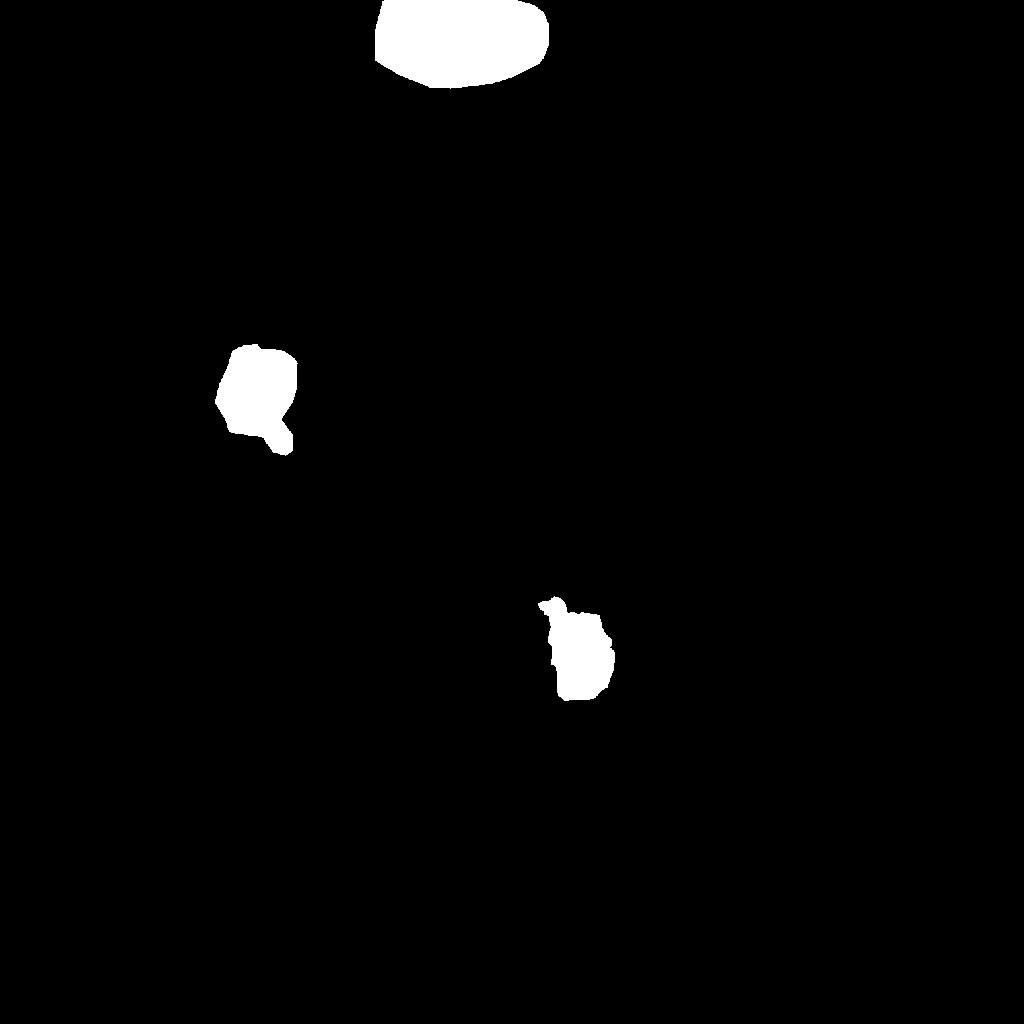} 
    \end{tabular}
    }
    \vspace{-8pt}
    \caption{\textbf{1-shot examples} from the MVTec dataset. Usually the anomaly is detected but the predicted anomaly area tends to be smaller than the ground-truth, hence the pixel-level ROC-AUC is smaller than the image-level.}
    \label{fig:mvtec_scores}
    \vspace{-15pt}
\end{figure}

\begin{figure*}[h!]
    \centering
    \resizebox{0.95\linewidth}{!}{
    \begin{tabular}{cccc|cccc}
    
        \multicolumn{4}{c|}{Normal} & \multicolumn{4}{c}{Anomaly} \\
        Query & Masked input & Recovered & Diff & Query & Masked input & Recovered & Diff \\
\includegraphics[width=0.125\linewidth]{figures/recovered_samples/bottle_n_orig.png} & 
\includegraphics[width=0.125\linewidth]{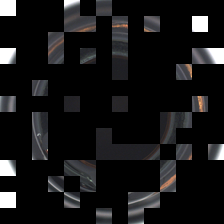} &
\includegraphics[width=0.125\linewidth]{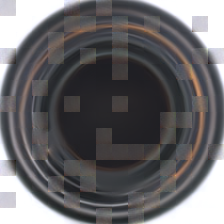} &
\includegraphics[width=0.125\linewidth]{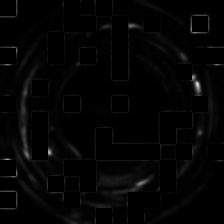} &
\includegraphics[width=0.125\linewidth]{figures/recovered_samples/bottle_a_orig.png} & 
\includegraphics[width=0.125\linewidth]{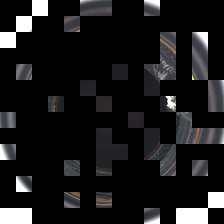} &
\includegraphics[width=0.125\linewidth]{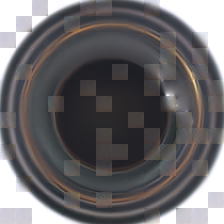} &
\includegraphics[width=0.125\linewidth]{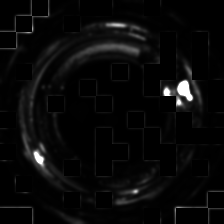}
\\
\includegraphics[width=0.125\linewidth]{figures/recovered_samples/capsule_n_orig.png} & 
\includegraphics[width=0.125\linewidth]{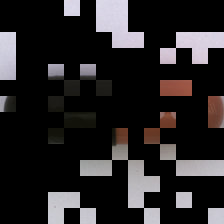} &
\includegraphics[width=0.125\linewidth]{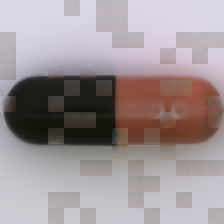} &
\includegraphics[width=0.125\linewidth]{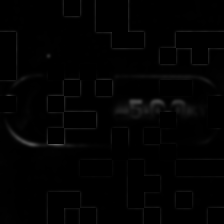} &
\includegraphics[width=0.125\linewidth]{figures/recovered_samples/capsule_a_orig.png} & 
\includegraphics[width=0.125\linewidth]{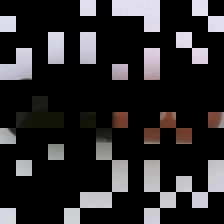} &
\includegraphics[width=0.125\linewidth]{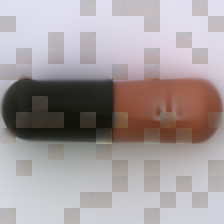} &
\includegraphics[width=0.125\linewidth]{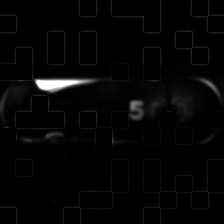}
\\
\includegraphics[width=0.125\linewidth]{figures/recovered_samples/carpet_n_orig.png} & 
\includegraphics[width=0.125\linewidth]{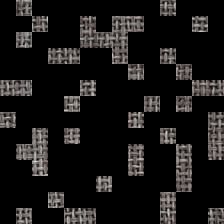} &
\includegraphics[width=0.125\linewidth]{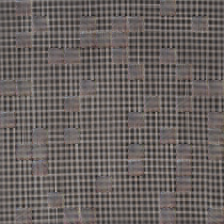} &
\includegraphics[width=0.125\linewidth]{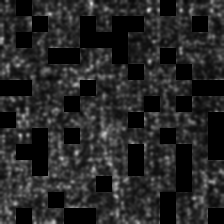} &
\includegraphics[width=0.125\linewidth]{figures/recovered_samples/carpet_a_orig.png} & 
\includegraphics[width=0.125\linewidth]{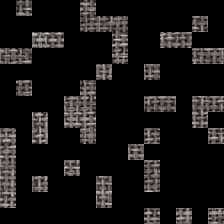} &
\includegraphics[width=0.125\linewidth]{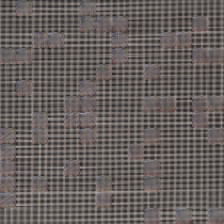} &
\includegraphics[width=0.125\linewidth]{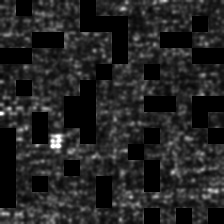}
\\
\includegraphics[width=0.125\linewidth]{figures/recovered_samples/grid_n_orig.png} & 
\includegraphics[width=0.125\linewidth]{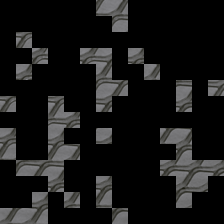} &
\includegraphics[width=0.125\linewidth]{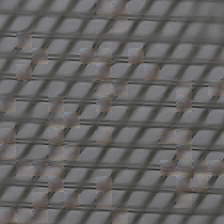} &
\includegraphics[width=0.125\linewidth]{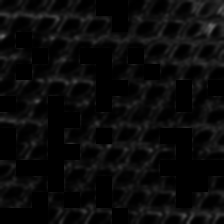} &
\includegraphics[width=0.125\linewidth]{figures/recovered_samples/grid_a_orig.png} & 
\includegraphics[width=0.125\linewidth]{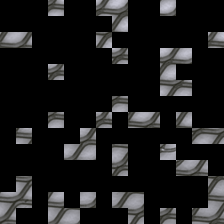} &
\includegraphics[width=0.125\linewidth]{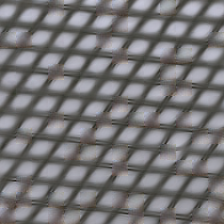} &
\includegraphics[width=0.125\linewidth]{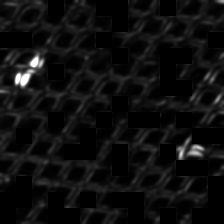}
\\
\includegraphics[width=0.125\linewidth]{figures/recovered_samples/hazelnut_n_orig.png} & 
\includegraphics[width=0.125\linewidth]{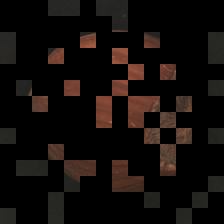} &
\includegraphics[width=0.125\linewidth]{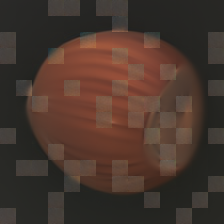} &
\includegraphics[width=0.125\linewidth]{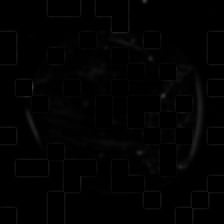} &
\includegraphics[width=0.125\linewidth]{figures/recovered_samples/hazelnut_a_orig.png} & 
\includegraphics[width=0.125\linewidth]{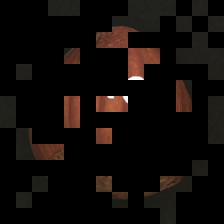} &
\includegraphics[width=0.125\linewidth]{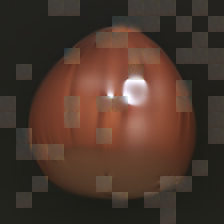} &
\includegraphics[width=0.125\linewidth]{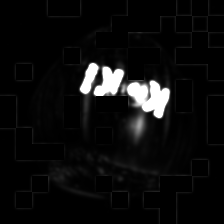}
\\
\includegraphics[width=0.125\linewidth]{figures/recovered_samples/leather_n_orig.png} & 
\includegraphics[width=0.125\linewidth]{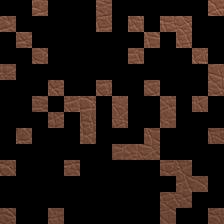} &
\includegraphics[width=0.125\linewidth]{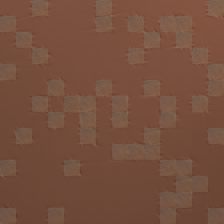} &
\includegraphics[width=0.125\linewidth]{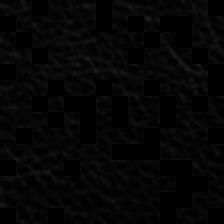} &
\includegraphics[width=0.125\linewidth]{figures/recovered_samples/leather_a_orig.png} & 
\includegraphics[width=0.125\linewidth]{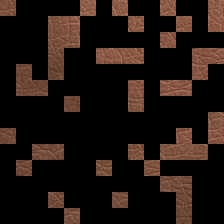} &
\includegraphics[width=0.125\linewidth]{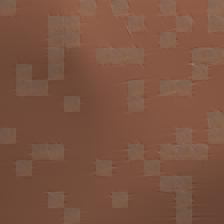} &
\includegraphics[width=0.125\linewidth]{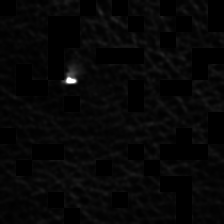}
\\
\includegraphics[width=0.125\linewidth]{figures/recovered_samples/pill_n_orig.png} & 
\includegraphics[width=0.125\linewidth]{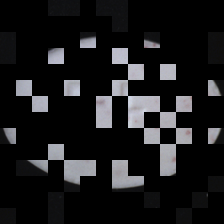} &
\includegraphics[width=0.125\linewidth]{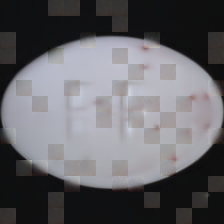} &
\includegraphics[width=0.125\linewidth]{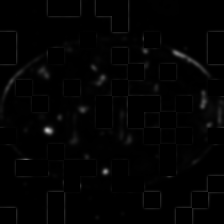} &
\includegraphics[width=0.125\linewidth]{figures/recovered_samples/pill_a_orig.png} & 
\includegraphics[width=0.125\linewidth]{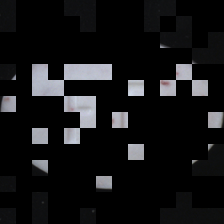} &
\includegraphics[width=0.125\linewidth]{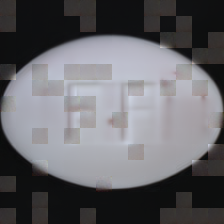} &
\includegraphics[width=0.125\linewidth]{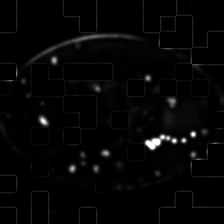}
\\
\includegraphics[width=0.125\linewidth]{figures/recovered_samples/screw_n_orig.png} & 
\includegraphics[width=0.125\linewidth]{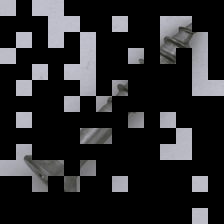} &
\includegraphics[width=0.125\linewidth]{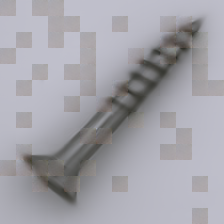} &
\includegraphics[width=0.125\linewidth]{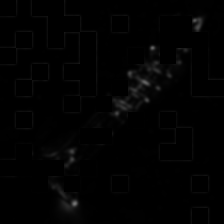} &
\includegraphics[width=0.125\linewidth]{figures/recovered_samples/screw_a_orig.png} & 
\includegraphics[width=0.125\linewidth]{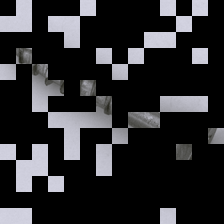} &
\includegraphics[width=0.125\linewidth]{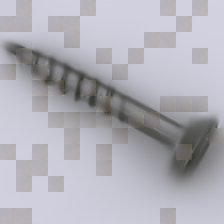} &
\includegraphics[width=0.125\linewidth]{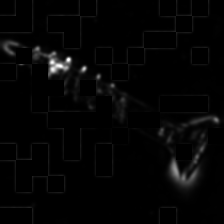}
\\
\includegraphics[width=0.125\linewidth]{figures/recovered_samples/tile_n_orig.png} & 
\includegraphics[width=0.125\linewidth]{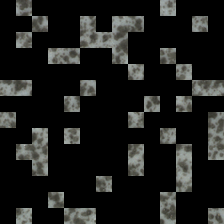} &
\includegraphics[width=0.125\linewidth]{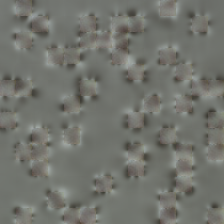} &
\includegraphics[width=0.125\linewidth]{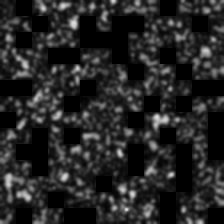} &
\includegraphics[width=0.125\linewidth]{figures/recovered_samples/tile_a_orig.png} & 
\includegraphics[width=0.125\linewidth]{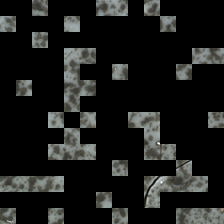} &
\includegraphics[width=0.125\linewidth]{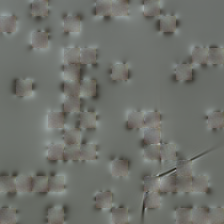} &
\includegraphics[width=0.125\linewidth]{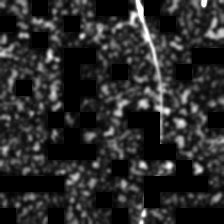}
\\
\includegraphics[width=0.125\linewidth]{figures/recovered_samples/wood_n_orig.png} & 
\includegraphics[width=0.125\linewidth]{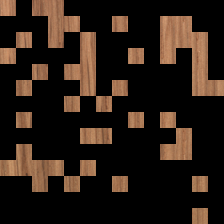} &
\includegraphics[width=0.125\linewidth]{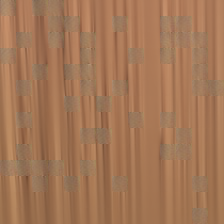} &
\includegraphics[width=0.125\linewidth]{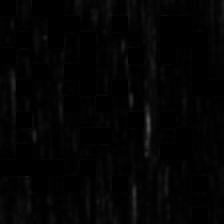} &
\includegraphics[width=0.125\linewidth]{figures/recovered_samples/wood_a_orig.png} & 
\includegraphics[width=0.125\linewidth]{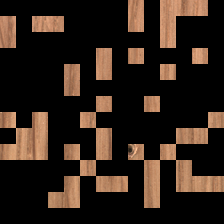} &
\includegraphics[width=0.125\linewidth]{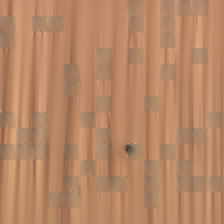} &
\includegraphics[width=0.125\linewidth]{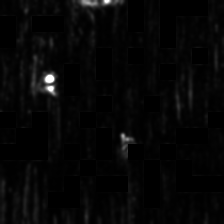}
\\
    \end{tabular}
    }
    \caption{Examples of reconstruction for both normal and anomalous images from the MVTech dataset. The model is usually able to recover (a blurry version of) the normal images. In many cases this is enough for detecting anomalous regions.}
    \label{fig:mvtec_recon}
\end{figure*}

\begin{figure*}[h!]
    \centering
    \resizebox{1\linewidth}{!}{
    \begin{tabular}{ccccc|ccccc}
    \multicolumn{5}{c|}{No Foreign Object} & \multicolumn{5}{c}{Foreign Object} \\
    Query & Masked & Recovered & Diff & Total Score & Query & Masked & Recovered & Diff & Total Score \\
    \includegraphics[width=0.1\linewidth, trim={40px 0px 40px 0px} ,clip]{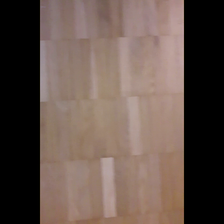} &
    \includegraphics[width=0.1\linewidth, trim={40px 0px 40px 0px} ,clip]{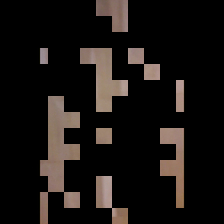} &
    \includegraphics[width=0.1\linewidth, trim={40px 0px 40px 0px} ,clip]{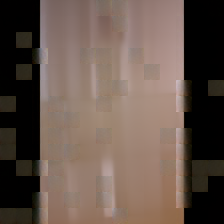} &
    \includegraphics[width=0.1\linewidth, trim={40px 0px 40px 0px} ,clip]{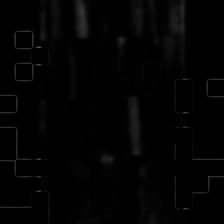} &
    \includegraphics[width=0.1\linewidth, trim={40px 0px 40px 0px} ,clip]{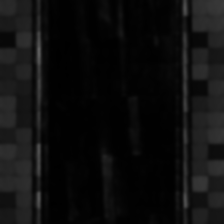}
    &
    \includegraphics[width=0.1\linewidth, trim={40px 0px 40px 0px} ,clip]{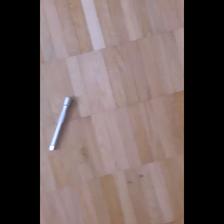} &
    \includegraphics[width=0.1\linewidth, trim={40px 0px 40px 0px} ,clip]{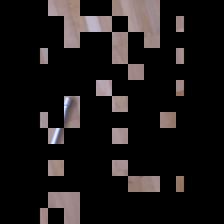} &
    \includegraphics[width=0.1\linewidth, trim={40px 0px 40px 0px} ,clip]{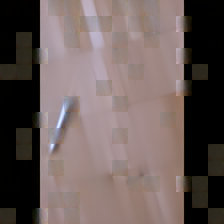} &
    \includegraphics[width=0.1\linewidth, trim={40px 0px 40px 0px} ,clip]{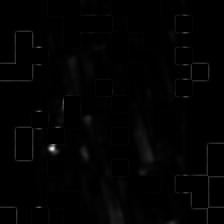} &
    \includegraphics[width=0.1\linewidth, trim={40px 0px 40px 0px} ,clip]{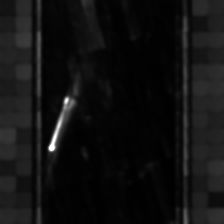}
    \\
    \includegraphics[width=0.1\linewidth, trim={40px 0px 40px 0px} ,clip]{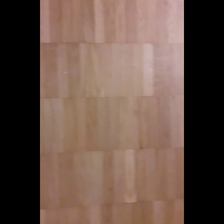} &
    \includegraphics[width=0.1\linewidth, trim={40px 0px 40px 0px} ,clip]{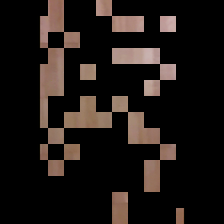} &
    \includegraphics[width=0.1\linewidth, trim={40px 0px 40px 0px} ,clip]{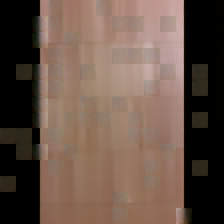} &
    \includegraphics[width=0.1\linewidth, trim={40px 0px 40px 0px} ,clip]{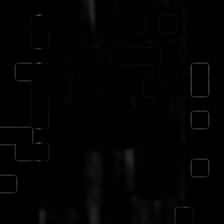} &
    \includegraphics[width=0.1\linewidth, trim={40px 0px 40px 0px} ,clip]{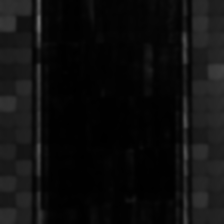}
    &
    \includegraphics[width=0.1\linewidth, trim={40px 0px 40px 0px} ,clip]{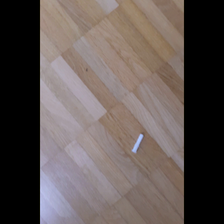} &
    \includegraphics[width=0.1\linewidth, trim={40px 0px 40px 0px} ,clip]{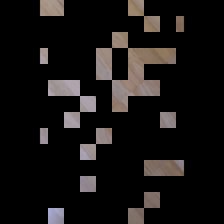} &
    \includegraphics[width=0.1\linewidth, trim={40px 0px 40px 0px} ,clip]{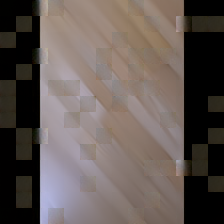} &
    \includegraphics[width=0.1\linewidth, trim={40px 0px 40px 0px} ,clip]{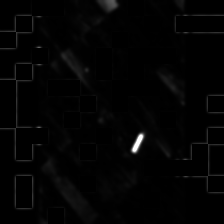} &
    \includegraphics[width=0.1\linewidth, trim={40px 0px 40px 0px} ,clip]{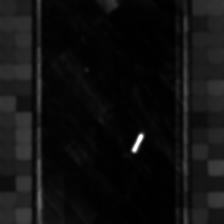}
    \\
    \includegraphics[width=0.1\linewidth, trim={40px 0px 40px 0px} ,clip]{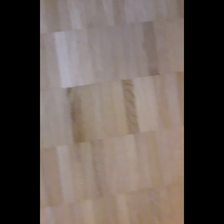} &
    \includegraphics[width=0.1\linewidth, trim={40px 0px 40px 0px} ,clip]{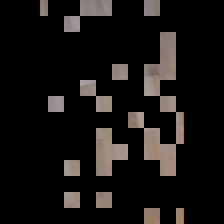} &
    \includegraphics[width=0.1\linewidth, trim={40px 0px 40px 0px} ,clip]{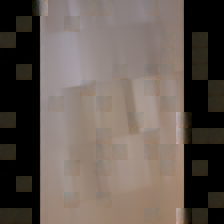} &
    \includegraphics[width=0.1\linewidth, trim={40px 0px 40px 0px} ,clip]{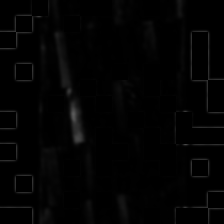} &
    \includegraphics[width=0.1\linewidth, trim={40px 0px 40px 0px} ,clip]{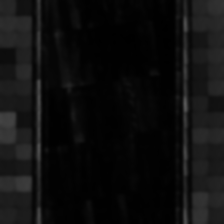}
    &
    \includegraphics[width=0.1\linewidth, trim={40px 0px 40px 0px} ,clip]{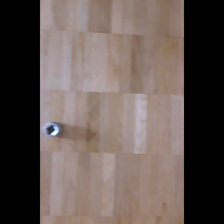} &
    \includegraphics[width=0.1\linewidth, trim={40px 0px 40px 0px} ,clip]{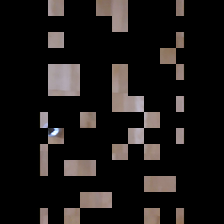} &
    \includegraphics[width=0.1\linewidth, trim={40px 0px 40px 0px} ,clip]{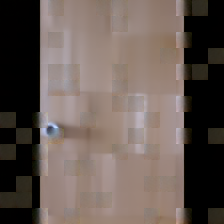} &
    \includegraphics[width=0.1\linewidth, trim={40px 0px 40px 0px} ,clip]{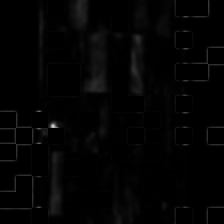} &
    \includegraphics[width=0.1\linewidth, trim={40px 0px 40px 0px} ,clip]{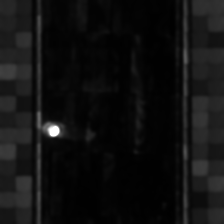}
    \\
    \includegraphics[width=0.1\linewidth, trim={40px 0px 40px 0px} ,clip]{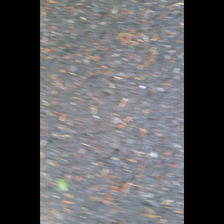} &
    \includegraphics[width=0.1\linewidth, trim={40px 0px 40px 0px} ,clip]{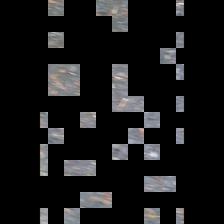} &
    \includegraphics[width=0.1\linewidth, trim={40px 0px 40px 0px} ,clip]{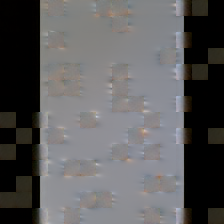} &
    \includegraphics[width=0.1\linewidth, trim={40px 0px 40px 0px} ,clip]{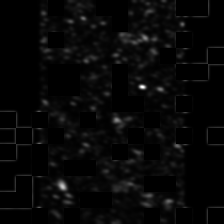} &
    \includegraphics[width=0.1\linewidth, trim={40px 0px 40px 0px} ,clip]{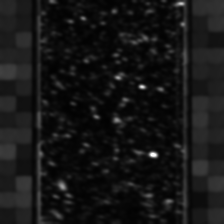}
    &
    \includegraphics[width=0.1\linewidth, trim={40px 0px 40px 0px} ,clip]{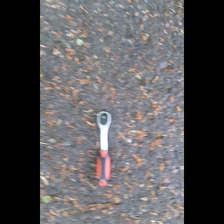} &
    \includegraphics[width=0.1\linewidth, trim={40px 0px 40px 0px} ,clip]{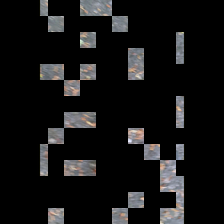} &
    \includegraphics[width=0.1\linewidth, trim={40px 0px 40px 0px} ,clip]{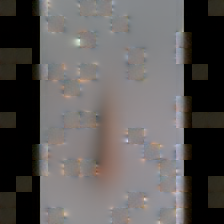} &
    \includegraphics[width=0.1\linewidth, trim={40px 0px 40px 0px} ,clip]{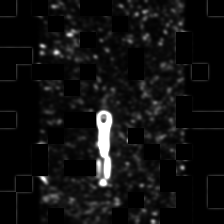} &
    \includegraphics[width=0.1\linewidth, trim={40px 0px 40px 0px} ,clip]{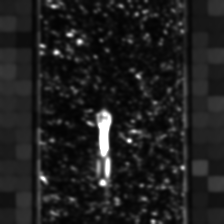}
    \\
    \includegraphics[width=0.1\linewidth, trim={40px 0px 40px 0px} ,clip]{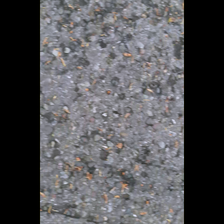} &
    \includegraphics[width=0.1\linewidth, trim={40px 0px 40px 0px} ,clip]{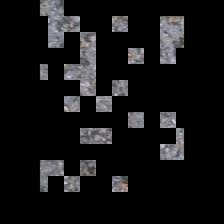} &
    \includegraphics[width=0.1\linewidth, trim={40px 0px 40px 0px} ,clip]{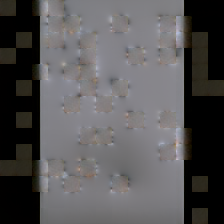} &
    \includegraphics[width=0.1\linewidth, trim={40px 0px 40px 0px} ,clip]{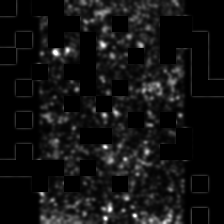} &
    \includegraphics[width=0.1\linewidth, trim={40px 0px 40px 0px} ,clip]{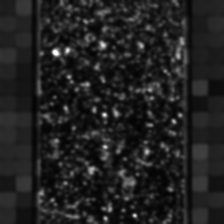}
    &
    \includegraphics[width=0.1\linewidth, trim={40px 0px 40px 0px} ,clip]{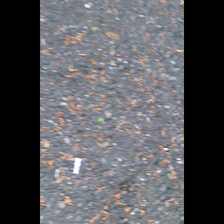} &
    \includegraphics[width=0.1\linewidth, trim={40px 0px 40px 0px} ,clip]{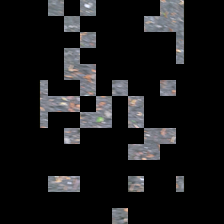} &
    \includegraphics[width=0.1\linewidth, trim={40px 0px 40px 0px} ,clip]{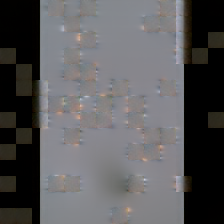} &
    \includegraphics[width=0.1\linewidth, trim={40px 0px 40px 0px} ,clip]{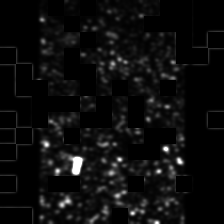} &
    \includegraphics[width=0.1\linewidth, trim={40px 0px 40px 0px} ,clip]{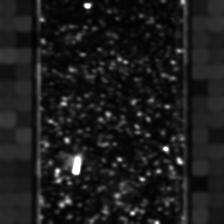}
    \\
    \includegraphics[width=0.1\linewidth, trim={40px 0px 40px 0px} ,clip]{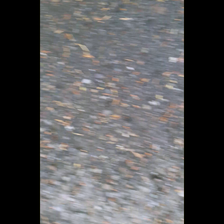} &
    \includegraphics[width=0.1\linewidth, trim={40px 0px 40px 0px} ,clip]{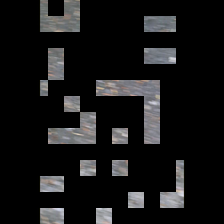} &
    \includegraphics[width=0.1\linewidth, trim={40px 0px 40px 0px} ,clip]{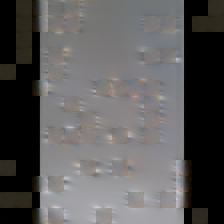} &
    \includegraphics[width=0.1\linewidth, trim={40px 0px 40px 0px} ,clip]{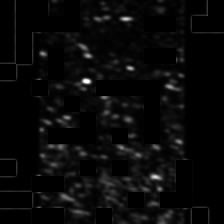} &
    \includegraphics[width=0.1\linewidth, trim={40px 0px 40px 0px} ,clip]{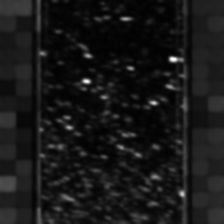}
    &
    \includegraphics[width=0.1\linewidth, trim={40px 0px 40px 0px} ,clip]{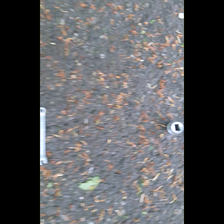} &
    \includegraphics[width=0.1\linewidth, trim={40px 0px 40px 0px} ,clip]{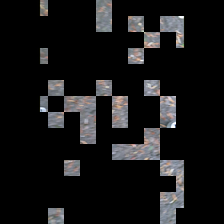} &
    \includegraphics[width=0.1\linewidth, trim={40px 0px 40px 0px} ,clip]{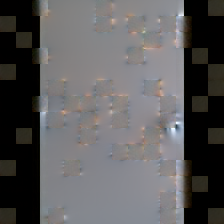} &
    \includegraphics[width=0.1\linewidth, trim={40px 0px 40px 0px} ,clip]{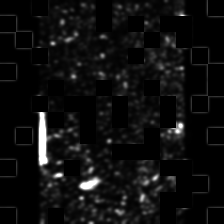} &
    \includegraphics[width=0.1\linewidth, trim={40px 0px 40px 0px} ,clip]{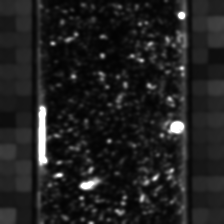}
    \\
    \end{tabular}
    }
    \caption{\textbf{ZSFOD \ours{} Foreign Object Detection results} with neither clean surface reference nor object references. ``Total Score" is the average of ``Diff" produced with $32$ different random masks applied to the same image. }
    \label{fig:fod_exaples}
\end{figure*}

We evaluated our method on all of the $15$ datasets in MVTec-AD \citep{bergmann2021mvtec}, the most popular and the main AD benchmark.
It is focused on an industrial inspection use case and consists of 10 unique objects and 5 unique textures. For each object or texture, a training set of defect-free images and a test set of both normal and anomalous instances are available. The anomalous images are provided with pixel-level annotation marking the anomaly location.

For the few-shot test, in each run, we selected a few random training samples from the relevant dataset training set and tested on the full associated test set. Since the performance can be dependent on the selected samples we averaged all results over 3 different shots selection. When comparing to other methods we made sure the same exact shots are used by all methods. When an ensemble of models is used, the same shots are used for all models, and the models' output images and pixel-level scores are summed. For the zero-shot test, per the task definition, the training set is not used.

Table \ref{tab:mvtec_img} summarizes the results for image-level zero and one-shot anomaly detection performance.
Even though for the zero-shot case \ours{} uses no normal training data, we observe relatively strong results.
For the textures datasets, it even outperforms the SOTA 1-shot results.
In the 1-shot case, we observe $1.5\%$ improvement of \ours{} thanks to the finetuning on the normal sample.
\resp{Notably, PatchCore is better than \ours{} on the objects subset while \ours{} is better on the textures subset. Suggesting our image-reconstruction-based method can better utilize the repetitive patterns as a self-reference compared to embedding-based methods.}
Next, we test the performance of an ensemble of two models.
While the ensemble of two PatchCore models outperforms a single one thanks to the stochastic nature of the method, the gain is limited by the fact they use the same embeddings and similar information.
The ensemble of \ours{} with PatchCore outperforms the two PatchCore ensemble by $3.1\%$.

Table \ref{tab:mvtec_pixel} summarizes the results for pixel-level zero and one-shot anomaly detection (segmentation) performance.
While the gap between \ours{} and PatchCore is higher for pixel-level detection, we observe similar trends to image-level performance.
For a single model, PatchCore outperforms \ours{}, but an ensemble of \ours{} and PatchCore is better than an ensemble of two PatchCore models.
We attribute the lower pixel-level performance to the fact that, even though \ours{} is mostly able to detect the anomalies, often the detected anomaly only partially covers the full anomaly region.
Examples of segmentation maps produced by \ours{} are presented in Figure \ref{fig:mvtec_scores}.
Examples of the recovered images from masked inputs are presented in Figure \ref{fig:mvtec_recon}, while the recovered images tend to be blurry they usually provide enough signal for detecting anomalies.

We explored finetuning \ours{} on more shots in Figure \ref{fig:more_shots}.
The improvement in the performance of \ours{} saturates at about 4 shots, making it a best fit for the low shot scenario.
For more shots, we observed similar results to 1-shot, where an ensemble of \ours{} and PatchCore sets a new SOTA.

\begin{table}[]
\caption{Training and inference time. Tested for 0/1 shot. \ours{} performs inference on a single image at a time to allow 32 repeats of the same image in the batch dimension (with different random masks). For PatchCore we used a batch size of 32. Despite that, the inference time is not dramatically higher for \ours{} compared to PatchCore. Tested on an A100 GPU.}
    \centering
    \begin{tabular}{lccc}
    \toprule
    & PatchCore & \ours{} & \ours{} \\
    & 1-shot &  0-shot & 1-shot \\
    \midrule
    Training &  4s & 0 & 100s \\
    Infer. [per image] & 0.07s & 0.15s & 0.15s\\ 
    \bottomrule
    \end{tabular}
    \label{tab:time}
\end{table}

\begin{table}[h]
    \caption{\textbf{LoRA ablation} Using LoRA for finetuning in a low-rank space improves performance.}
    \centering
    \begin{tabular}{lcc}
    \toprule
    & Full-finetuning & LoRA finetuning\\
    \midrule
    Image ROC-AUC &  75.4  & 76.0 \\
    Pixel ROC-AUC   &  70.9 & 71.6 \\
         \bottomrule
    \end{tabular}
    \label{tab:lora}
\end{table}

\begin{table}[]
    \centering
        \caption{\textbf{Foreign Object Detection} ROC-AUC performance for zero-shot Foreign Object Detection (ZSFOD). \ours{}, which is a 0-shot method, outperforms the 1-shot AD baseline.}
    \begin{tabular}{lccc|c}
    \toprule
        Method & Shots & Indoor & Outdoor & Mean \\
        \midrule
        PatchCore & 1 & \textbf{98.2} & 73.2 & 85.7 \\ 
        \ours{} & 0 & 95.6 & \textbf{85.6} & \textbf{90.6} \\
        \bottomrule
    \end{tabular}
    \label{tab:fod}
\end{table}

\paragraph{Num. of repetitions per image}
We tested the effect of averaging the anomaly score from multiple inferences of the same image with different random masks.
We used a varying number of repetitions per image, from as little as a single run to 64 runs.
In Figure \ref{fig:num_reps} we summarized the results.
The performance seems to saturate at $\sim32$ repetitions.

\paragraph{LoRA}
In Table \ref{tab:lora} we compare the performance of finetuning the original model parameters vs. training a low-rank version of them using LoRA.
For finetuning without LoRA we used a learning rate of $1e-4$ with all other hyperparameters unchanged. We observe $0.6\%$ improvement in image-level performance and $0.8\%$ in pixel-level performance when using LoRA.

\paragraph{Training and inference time}
We compare training and inference time in Table \ref{tab:time}.
We tested the running time for the 0-shot and 1-shot cases.
Time was measured on an A100 GPU.
For PatchCore training includes extracting features using a pretrained model and performing CoreSet clustering and amounts for 4 seconds.
The reported training time for \ours{} was measured when training for 50 iterations and took around 100 seconds.
There is a trade-off between finetuning \ours{} which takes time and using \ours{} in its 0-shot form which does not require training but with an accuracy drop of $1.5\%$.
The inference was performed in batches for PatchCore (batch-size=$32$). For \ours{}, each query image is processed individually since we use the batch dimension to run multiple instances of the same image with different random masks.
Despite the parallelization in PatchCore (and the lack of it in \ours{}) the inference time is in the same order of magnitude with 0.07 seconds for PatchCore and 0.15 for \ours{}.
This is partially thanks to the fact the MAE's encoder inputs are only $25\%$ of the tokens. The $75\%$ of the tokens that need to be reconstructed are only introduced later as inputs to the decoder which is a much smaller network.

\subsection{Foreign Object Detection}
We also tested a proof-of-concept of using \ours{} for Zero-Shot Foreign Object Detection (ZSFOD).
FOD is a very important task in several real-world scenarios, e.g. in airport runways, where even very small objects on the ground can be dangerous for the planes.
Unlike classic FOD where models are trained for detecting specific types of objects, here no training data of either an empty surface or the objects to be detected are provided.
We treat FOD as detecting anomalies in the background surface texture.
We captured videos of the ground in two environments, indoors (wooden floor) and outdoors (asphalt pavement). Some of the frames contain foreign objects. Objects include larger tools, e.g. a wrench, and smaller objects, e.g. a bolt. We extracted and labeled 20-50 frames with foreign objects and a similar number without any object for each of the environments.
This dataset will be released.

Since we are the first to perform the task of ZSFOD, we chose to compare \ourss against the SOTA 1-shot AD method, PatchCore \citep{roth2022towards}. This is a very strong baseline since it uses an object-free reference.
Table \ref{tab:fod} summarizes the results.
\resp{
We observed strong results by \ours{} for ZSFOD, \ourss performs close to (Indoors) or better (Outdoors) compared to 1-shot PatchCore, with an average improvement of $+4.9\%$.
}
Examples of images from the dataset along with their recovered outputs by \ours{} and the final segmentation results are presented in Figure~\ref{fig:fod_exaples}.

\section{Conclusions and Future Work}
We have suggested \ours{}, using an ImageNet pretrained MAE for the task of few-shot anomaly detection (FSAD).
This is the first image-reconstruction-based method used in the low-shot regime.
While image-reconstruction-based methods are not the strongest methods for AD, we showed they provide additional valuable information.
An ensemble of an embedding-based method and \ours{} sets a new SOTA for FSAD.

We have also suggested the new Zero-Shot Anomaly-Detection task (ZSAD), performing anomaly detection with no reference images. 
We have shown \ours{} can be used for this task and performs surprisingly well despite working with novel objects and textures.
Specifically for textures, \ours{} outperforms the reference-based FSAD SOTA baseline.

We explored a new task of Foreign Object Detection (FOD) on the ground, with no prior reference to either a free-of-objects surface or to the objects to be detected.
We treated this problem as ZSAD where the objects are an anomaly in the surface texture.
We showed better results for this task compared with SOTA FSAD where an image of the surface is provided for reference. The dataset is also made available to the community.

In future work \ours{} can be extended to better use the few available shots in FSAD.
We can feed the model tokens (patches) from both the query image and the reference image(s).
The model can be trained to use the transformer's attention mechanism to share information between the reference tokens and query tokens.
This way the recovered patches are not just guessed according to their surrounding and are more likely to fit the normal patch distribution.

\section*{Acknowledgments}
This work was supported by the European Research Council
(ERC-StG 757497 PI Giryes).

\clearpage
\bibliographystyle{model2-names}
\bibliography{refs}

\begin{thebibliography}{36}
\expandafter\ifx\csname natexlab\endcsname\relax\def\natexlab#1{#1}\fi
\providecommand{\url}[1]{\texttt{#1}}
\providecommand{\href}[2]{#2}
\providecommand{\path}[1]{#1}
\providecommand{\DOIprefix}{doi:}
\providecommand{\ArXivprefix}{arXiv:}
\providecommand{\URLprefix}{URL: }
\providecommand{\Pubmedprefix}{pmid:}
\providecommand{\doi}[1]{\href{http://dx.doi.org/#1}{\path{#1}}}
\providecommand{\Pubmed}[1]{\href{pmid:#1}{\path{#1}}}
\providecommand{\bibinfo}[2]{#2}
\ifx\xfnm\relax \def\xfnm[#1]{\unskip,\space#1}\fi
\bibitem[{Bergmann et~al.(2021)Bergmann, Batzner, Fauser, Sattlegger and
  Steger}]{bergmann2021mvtec}
\bibinfo{author}{Bergmann, P.}, \bibinfo{author}{Batzner, K.},
  \bibinfo{author}{Fauser, M.}, \bibinfo{author}{Sattlegger, D.},
  \bibinfo{author}{Steger, C.}, \bibinfo{year}{2021}.
\newblock \bibinfo{title}{The mvtec anomaly detection dataset: a comprehensive
  real-world dataset for unsupervised anomaly detection}.
\newblock \bibinfo{journal}{International Journal of Computer Vision}
  \bibinfo{volume}{129}, \bibinfo{pages}{1038--1059}.
\bibitem[{Cao et~al.(2023)Cao, Xu, Sun, Cheng, Du, Gao and
  Shen}]{cao2023segment}
\bibinfo{author}{Cao, Y.}, \bibinfo{author}{Xu, X.}, \bibinfo{author}{Sun, C.},
  \bibinfo{author}{Cheng, Y.}, \bibinfo{author}{Du, Z.}, \bibinfo{author}{Gao,
  L.}, \bibinfo{author}{Shen, W.}, \bibinfo{year}{2023}.
\newblock \bibinfo{title}{Segment any anomaly without training via hybrid
  prompt regularization}.
\newblock \bibinfo{journal}{arXiv preprint} .
\bibitem[{Chen et~al.(1999)Chen, Kanade, Pomerleau and
  Rowley}]{chen1999anomaly}
\bibinfo{author}{Chen, M.}, \bibinfo{author}{Kanade, T.},
  \bibinfo{author}{Pomerleau, D.}, \bibinfo{author}{Rowley, H.A.},
  \bibinfo{year}{1999}.
\newblock \bibinfo{title}{Anomaly detection through registration}.
\newblock \bibinfo{journal}{Pattern Recognition} \bibinfo{volume}{32},
  \bibinfo{pages}{113--128}.
\bibitem[{Cohen and Hoshen(2020)}]{cohen2020sub}
\bibinfo{author}{Cohen, N.}, \bibinfo{author}{Hoshen, Y.},
  \bibinfo{year}{2020}.
\newblock \bibinfo{title}{Sub-image anomaly detection with deep pyramid
  correspondences}.
\newblock \bibinfo{journal}{arXiv preprint arXiv:2005.02357} .
\bibitem[{Defard et~al.(2021)Defard, Setkov, Loesch and
  Audigier}]{defard2021padim}
\bibinfo{author}{Defard, T.}, \bibinfo{author}{Setkov, A.},
  \bibinfo{author}{Loesch, A.}, \bibinfo{author}{Audigier, R.},
  \bibinfo{year}{2021}.
\newblock \bibinfo{title}{Padim: a patch distribution modeling framework for
  anomaly detection and localization}, in: \bibinfo{booktitle}{International
  Conference on Pattern Recognition}, \bibinfo{organization}{Springer}. pp.
  \bibinfo{pages}{475--489}.
\bibitem[{Doveh et~al.(2021)Doveh, Schwartz, Xue, Feris, Bronstein, Giryes and
  Karlinsky}]{doveh2021metadapt}
\bibinfo{author}{Doveh, S.}, \bibinfo{author}{Schwartz, E.},
  \bibinfo{author}{Xue, C.}, \bibinfo{author}{Feris, R.},
  \bibinfo{author}{Bronstein, A.}, \bibinfo{author}{Giryes, R.},
  \bibinfo{author}{Karlinsky, L.}, \bibinfo{year}{2021}.
\newblock \bibinfo{title}{Metadapt: Meta-learned task-adaptive architecture for
  few-shot classification}.
\newblock \bibinfo{journal}{Pattern Recognition Letters} \bibinfo{volume}{149},
  \bibinfo{pages}{130--136}.
\bibitem[{Fei et~al.(2020)Fei, Huang, Jinkun, Li, Zhang and
  Lu}]{fei2020attribute}
\bibinfo{author}{Fei, Y.}, \bibinfo{author}{Huang, C.},
  \bibinfo{author}{Jinkun, C.}, \bibinfo{author}{Li, M.},
  \bibinfo{author}{Zhang, Y.}, \bibinfo{author}{Lu, C.}, \bibinfo{year}{2020}.
\newblock \bibinfo{title}{Attribute restoration framework for anomaly
  detection}.
\newblock \bibinfo{journal}{IEEE Transactions on Multimedia} .
\bibitem[{Goodfellow et~al.(2014)Goodfellow, Pouget-Abadie, Mirza, Xu,
  Warde-Farley, Ozair, Courville and Bengio}]{goodfellow2014generative}
\bibinfo{author}{Goodfellow, I.}, \bibinfo{author}{Pouget-Abadie, J.},
  \bibinfo{author}{Mirza, M.}, \bibinfo{author}{Xu, B.},
  \bibinfo{author}{Warde-Farley, D.}, \bibinfo{author}{Ozair, S.},
  \bibinfo{author}{Courville, A.}, \bibinfo{author}{Bengio, Y.},
  \bibinfo{year}{2014}.
\newblock \bibinfo{title}{Generative adversarial nets}.
\newblock \bibinfo{journal}{Advances in neural information processing systems}
  \bibinfo{volume}{27}.
\bibitem[{Gudovskiy et~al.(2022)Gudovskiy, Ishizaka and
  Kozuka}]{gudovskiy2022cflow}
\bibinfo{author}{Gudovskiy, D.}, \bibinfo{author}{Ishizaka, S.},
  \bibinfo{author}{Kozuka, K.}, \bibinfo{year}{2022}.
\newblock \bibinfo{title}{Cflow-ad: Real-time unsupervised anomaly detection
  with localization via conditional normalizing flows}, in:
  \bibinfo{booktitle}{Proceedings of the IEEE/CVF Winter Conference on
  Applications of Computer Vision}, pp. \bibinfo{pages}{98--107}.
\bibitem[{He et~al.(2022)He, Chen, Xie, Li, Doll{\'a}r and
  Girshick}]{he2022masked}
\bibinfo{author}{He, K.}, \bibinfo{author}{Chen, X.}, \bibinfo{author}{Xie,
  S.}, \bibinfo{author}{Li, Y.}, \bibinfo{author}{Doll{\'a}r, P.},
  \bibinfo{author}{Girshick, R.}, \bibinfo{year}{2022}.
\newblock \bibinfo{title}{Masked autoencoders are scalable vision learners},
  in: \bibinfo{booktitle}{Proceedings of the IEEE/CVF Conference on Computer
  Vision and Pattern Recognition}, pp. \bibinfo{pages}{16000--16009}.
\bibitem[{Hinton(1990)}]{hinton1990connectionist}
\bibinfo{author}{Hinton, G.E.}, \bibinfo{year}{1990}.
\newblock \bibinfo{title}{Connectionist learning procedures}, in:
  \bibinfo{booktitle}{Machine learning}. \bibinfo{publisher}{Elsevier}, pp.
  \bibinfo{pages}{555--610}.
\bibitem[{Hu et~al.(2021)Hu, Shen, Wallis, Allen-Zhu, Li, Wang, Wang and
  Chen}]{hu2021lora}
\bibinfo{author}{Hu, E.J.}, \bibinfo{author}{Shen, Y.},
  \bibinfo{author}{Wallis, P.}, \bibinfo{author}{Allen-Zhu, Z.},
  \bibinfo{author}{Li, Y.}, \bibinfo{author}{Wang, S.}, \bibinfo{author}{Wang,
  L.}, \bibinfo{author}{Chen, W.}, \bibinfo{year}{2021}.
\newblock \bibinfo{title}{Lora: Low-rank adaptation of large language models}.
\newblock \bibinfo{journal}{arXiv preprint arXiv:2106.09685} .
\bibitem[{Huang et~al.(2022)Huang, Guan, Jiang, Zhang, Spratling and
  Wang}]{huang2022registration}
\bibinfo{author}{Huang, C.}, \bibinfo{author}{Guan, H.},
  \bibinfo{author}{Jiang, A.}, \bibinfo{author}{Zhang, Y.},
  \bibinfo{author}{Spratling, M.}, \bibinfo{author}{Wang, Y.F.},
  \bibinfo{year}{2022}.
\newblock \bibinfo{title}{Registration based few-shot anomaly detection}, in:
  \bibinfo{booktitle}{European Conference on Computer Vision},
  \bibinfo{organization}{Springer}. pp. \bibinfo{pages}{303--319}.
\bibitem[{Japkowicz et~al.(1995)Japkowicz, Myers, Gluck
  et~al.}]{japkowicz1995novelty}
\bibinfo{author}{Japkowicz, N.}, \bibinfo{author}{Myers, C.},
  \bibinfo{author}{Gluck, M.}, et~al., \bibinfo{year}{1995}.
\newblock \bibinfo{title}{A novelty detection approach to classification}, in:
  \bibinfo{booktitle}{IJCAI}, \bibinfo{organization}{Citeseer}. pp.
  \bibinfo{pages}{518--523}.
\bibitem[{Jeong et~al.(2023)Jeong, Zou, Kim, Zhang, Ravichandran and
  Dabeer}]{jeong2023winclip}
\bibinfo{author}{Jeong, J.}, \bibinfo{author}{Zou, Y.}, \bibinfo{author}{Kim,
  T.}, \bibinfo{author}{Zhang, D.}, \bibinfo{author}{Ravichandran, A.},
  \bibinfo{author}{Dabeer, O.}, \bibinfo{year}{2023}.
\newblock \bibinfo{title}{Winclip: Zero-\/few-shot anomaly classification and
  segmentation}, in: \bibinfo{booktitle}{Proceedings of the IEEE Conference on
  Computer Vision and Pattern Recognition (CVPR)}, pp.
  \bibinfo{pages}{19606--19616}.
\bibitem[{Jing et~al.(2022)Jing, Zheng, Zheng and Dong}]{jing2022pixel}
\bibinfo{author}{Jing, Y.}, \bibinfo{author}{Zheng, H.},
  \bibinfo{author}{Zheng, W.}, \bibinfo{author}{Dong, K.},
  \bibinfo{year}{2022}.
\newblock \bibinfo{title}{A pixel-wise foreign object debris detection method
  based on multi-scale feature inpainting}.
\newblock \bibinfo{journal}{Aerospace} \bibinfo{volume}{9},
  \bibinfo{pages}{480}.
\bibitem[{Li et~al.(2021)Li, Sohn, Yoon and Pfister}]{li2021cutpaste}
\bibinfo{author}{Li, C.L.}, \bibinfo{author}{Sohn, K.}, \bibinfo{author}{Yoon,
  J.}, \bibinfo{author}{Pfister, T.}, \bibinfo{year}{2021}.
\newblock \bibinfo{title}{Cutpaste: Self-supervised learning for anomaly
  detection and localization}, in: \bibinfo{booktitle}{Proceedings of the
  IEEE/CVF Conference on Computer Vision and Pattern Recognition}, pp.
  \bibinfo{pages}{9664--9674}.
\bibitem[{Munyer et~al.(2022)Munyer, Brinkman, Zhong, Huang and
  Konstantzos}]{munyer2022foreign}
\bibinfo{author}{Munyer, T.}, \bibinfo{author}{Brinkman, D.},
  \bibinfo{author}{Zhong, X.}, \bibinfo{author}{Huang, C.},
  \bibinfo{author}{Konstantzos, I.}, \bibinfo{year}{2022}.
\newblock \bibinfo{title}{Foreign object debris detection for airport pavement
  images based on self-supervised localization and vision transformer}.
\newblock \bibinfo{journal}{arXiv preprint arXiv:2210.16901} .
\bibitem[{Munyer et~al.(2021)Munyer, Huang, Huang and Zhong}]{munyer2021fod}
\bibinfo{author}{Munyer, T.}, \bibinfo{author}{Huang, P.C.},
  \bibinfo{author}{Huang, C.}, \bibinfo{author}{Zhong, X.},
  \bibinfo{year}{2021}.
\newblock \bibinfo{title}{Fod-a: A dataset for foreign object debris in
  airports}.
\newblock \bibinfo{journal}{arXiv preprint arXiv:2110.03072} .
\bibitem[{Noroozi and Shah(2023)}]{noroozi2023towards}
\bibinfo{author}{Noroozi, M.}, \bibinfo{author}{Shah, A.},
  \bibinfo{year}{2023}.
\newblock \bibinfo{title}{Towards optimal foreign object debris detection in an
  airport environment}.
\newblock \bibinfo{journal}{Expert Systems with Applications}
  \bibinfo{volume}{213}, \bibinfo{pages}{118829}.
\bibitem[{Rezende and Mohamed(2015)}]{rezende2015variational}
\bibinfo{author}{Rezende, D.}, \bibinfo{author}{Mohamed, S.},
  \bibinfo{year}{2015}.
\newblock \bibinfo{title}{Variational inference with normalizing flows}, in:
  \bibinfo{booktitle}{International conference on machine learning},
  \bibinfo{organization}{PMLR}. pp. \bibinfo{pages}{1530--1538}.
\bibitem[{Roth et~al.(2022)Roth, Pemula, Zepeda, Sch{\"o}lkopf, Brox and
  Gehler}]{roth2022towards}
\bibinfo{author}{Roth, K.}, \bibinfo{author}{Pemula, L.},
  \bibinfo{author}{Zepeda, J.}, \bibinfo{author}{Sch{\"o}lkopf, B.},
  \bibinfo{author}{Brox, T.}, \bibinfo{author}{Gehler, P.},
  \bibinfo{year}{2022}.
\newblock \bibinfo{title}{Towards total recall in industrial anomaly
  detection}, in: \bibinfo{booktitle}{Proceedings of the IEEE/CVF Conference on
  Computer Vision and Pattern Recognition}, pp. \bibinfo{pages}{14318--14328}.
\bibitem[{Rudolph et~al.(2021)Rudolph, Wandt and Rosenhahn}]{Rudolph_2021_WACV}
\bibinfo{author}{Rudolph, M.}, \bibinfo{author}{Wandt, B.},
  \bibinfo{author}{Rosenhahn, B.}, \bibinfo{year}{2021}.
\newblock \bibinfo{title}{Same same but differnet: Semi-supervised defect
  detection with normalizing flows}, in: \bibinfo{booktitle}{Proceedings of the
  IEEE/CVF Winter Conference on Applications of Computer Vision (WACV)}, pp.
  \bibinfo{pages}{1907--1916}.
\bibitem[{Sakurada and Yairi(2014)}]{sakurada2014anomaly}
\bibinfo{author}{Sakurada, M.}, \bibinfo{author}{Yairi, T.},
  \bibinfo{year}{2014}.
\newblock \bibinfo{title}{Anomaly detection using autoencoders with nonlinear
  dimensionality reduction}, in: \bibinfo{booktitle}{Proceedings of the MLSDA
  2014 2nd workshop on machine learning for sensory data analysis}, pp.
  \bibinfo{pages}{4--11}.
\bibitem[{Schlegl et~al.(2017)Schlegl, Seeb{\"o}ck, Waldstein, Schmidt-Erfurth
  and Langs}]{schlegl2017unsupervised}
\bibinfo{author}{Schlegl, T.}, \bibinfo{author}{Seeb{\"o}ck, P.},
  \bibinfo{author}{Waldstein, S.M.}, \bibinfo{author}{Schmidt-Erfurth, U.},
  \bibinfo{author}{Langs, G.}, \bibinfo{year}{2017}.
\newblock \bibinfo{title}{Unsupervised anomaly detection with generative
  adversarial networks to guide marker discovery}, in:
  \bibinfo{booktitle}{International conference on information processing in
  medical imaging}, \bibinfo{organization}{Springer}. pp.
  \bibinfo{pages}{146--157}.
\bibitem[{Sheynin et~al.(2021)Sheynin, Benaim and Wolf}]{Sheynin_2021_ICCV}
\bibinfo{author}{Sheynin, S.}, \bibinfo{author}{Benaim, S.},
  \bibinfo{author}{Wolf, L.}, \bibinfo{year}{2021}.
\newblock \bibinfo{title}{A hierarchical transformation-discriminating
  generative model for few shot anomaly detection}, in:
  \bibinfo{booktitle}{Proceedings of the IEEE/CVF International Conference on
  Computer Vision (ICCV)}, pp. \bibinfo{pages}{8495--8504}.
\bibitem[{Snell et~al.(2017)Snell, Swersky and Zemel}]{snell2017prototypical}
\bibinfo{author}{Snell, J.}, \bibinfo{author}{Swersky, K.},
  \bibinfo{author}{Zemel, R.}, \bibinfo{year}{2017}.
\newblock \bibinfo{title}{Prototypical networks for few-shot learning}.
\newblock \bibinfo{journal}{Advances in neural information processing systems}
  \bibinfo{volume}{30}.
\bibitem[{Vinyals et~al.(2016)Vinyals, Blundell, Lillicrap, Wierstra
  et~al.}]{vinyals2016matching}
\bibinfo{author}{Vinyals, O.}, \bibinfo{author}{Blundell, C.},
  \bibinfo{author}{Lillicrap, T.}, \bibinfo{author}{Wierstra, D.}, et~al.,
  \bibinfo{year}{2016}.
\newblock \bibinfo{title}{Matching networks for one shot learning}.
\newblock \bibinfo{journal}{Advances in neural information processing systems}
  \bibinfo{volume}{29}.
\bibitem[{Wyatt et~al.(2022)Wyatt, Leach, Schmon and
  Willcocks}]{wyatt2022anoddpm}
\bibinfo{author}{Wyatt, J.}, \bibinfo{author}{Leach, A.},
  \bibinfo{author}{Schmon, S.M.}, \bibinfo{author}{Willcocks, C.G.},
  \bibinfo{year}{2022}.
\newblock \bibinfo{title}{Anoddpm: Anomaly detection with denoising diffusion
  probabilistic models using simplex noise}, in:
  \bibinfo{booktitle}{Proceedings of the IEEE/CVF Conference on Computer Vision
  and Pattern Recognition}, pp. \bibinfo{pages}{650--656}.
\bibitem[{Xia et~al.(2022)Xia, Pan, Li, He, Ma, Zhang and Ding}]{xia2022gan}
\bibinfo{author}{Xia, X.}, \bibinfo{author}{Pan, X.}, \bibinfo{author}{Li, N.},
  \bibinfo{author}{He, X.}, \bibinfo{author}{Ma, L.}, \bibinfo{author}{Zhang,
  X.}, \bibinfo{author}{Ding, N.}, \bibinfo{year}{2022}.
\newblock \bibinfo{title}{Gan-based anomaly detection: A review}.
\newblock \bibinfo{journal}{Neurocomputing} .
\bibitem[{Yan et~al.(2021)Yan, Zhang, Xu, Hu and Heng}]{yan2021learning}
\bibinfo{author}{Yan, X.}, \bibinfo{author}{Zhang, H.}, \bibinfo{author}{Xu,
  X.}, \bibinfo{author}{Hu, X.}, \bibinfo{author}{Heng, P.A.},
  \bibinfo{year}{2021}.
\newblock \bibinfo{title}{Learning semantic context from normal samples for
  unsupervised anomaly detection}, in: \bibinfo{booktitle}{Proceedings of the
  AAAI Conference on Artificial Intelligence}, pp. \bibinfo{pages}{3110--3118}.
\bibitem[{Yu et~al.(2021)Yu, Zheng, Wang, Li, Wu, Zhao and Wu}]{yu2021fastflow}
\bibinfo{author}{Yu, J.}, \bibinfo{author}{Zheng, Y.}, \bibinfo{author}{Wang,
  X.}, \bibinfo{author}{Li, W.}, \bibinfo{author}{Wu, Y.},
  \bibinfo{author}{Zhao, R.}, \bibinfo{author}{Wu, L.}, \bibinfo{year}{2021}.
\newblock \bibinfo{title}{Fastflow: Unsupervised anomaly detection and
  localization via 2d normalizing flows}.
\newblock \bibinfo{journal}{arXiv preprint arXiv:2111.07677} .
\bibitem[{Zavrtanik et~al.(2021)Zavrtanik, Kristan and
  Sko{\v{c}}aj}]{zavrtanik2021reconstruction}
\bibinfo{author}{Zavrtanik, V.}, \bibinfo{author}{Kristan, M.},
  \bibinfo{author}{Sko{\v{c}}aj, D.}, \bibinfo{year}{2021}.
\newblock \bibinfo{title}{Reconstruction by inpainting for visual anomaly
  detection}.
\newblock \bibinfo{journal}{Pattern Recognition} \bibinfo{volume}{112},
  \bibinfo{pages}{107706}.
\bibitem[{Zenati et~al.(2018)Zenati, Foo, Lecouat, Manek and
  Chandrasekhar}]{zenati2018efficient}
\bibinfo{author}{Zenati, H.}, \bibinfo{author}{Foo, C.S.},
  \bibinfo{author}{Lecouat, B.}, \bibinfo{author}{Manek, G.},
  \bibinfo{author}{Chandrasekhar, V.R.}, \bibinfo{year}{2018}.
\newblock \bibinfo{title}{Efficient gan-based anomaly detection}.
\newblock \bibinfo{journal}{arXiv preprint arXiv:1802.06222} .
\bibitem[{Zhang et~al.(2021)Zhang, Saleeby, Feldhausen, Bi, Plotkowski and
  Womble}]{zhang2021self}
\bibinfo{author}{Zhang, J.}, \bibinfo{author}{Saleeby, K.},
  \bibinfo{author}{Feldhausen, T.}, \bibinfo{author}{Bi, S.},
  \bibinfo{author}{Plotkowski, A.}, \bibinfo{author}{Womble, D.},
  \bibinfo{year}{2021}.
\newblock \bibinfo{title}{Self-supervised anomaly detection via neural
  autoregressive flows with active learning}, in: \bibinfo{booktitle}{NeurIPS
  2021 Workshop on Deep Generative Models and Downstream Applications}.
\bibitem[{Zou et~al.(2022)Zou, Jeong, Pemula, Zhang and Dabeer}]{zou2022spot}
\bibinfo{author}{Zou, Y.}, \bibinfo{author}{Jeong, J.},
  \bibinfo{author}{Pemula, L.}, \bibinfo{author}{Zhang, D.},
  \bibinfo{author}{Dabeer, O.}, \bibinfo{year}{2022}.
\newblock \bibinfo{title}{Spot-the-difference self-supervised pre-training for
  anomaly detection and segmentation}, in: \bibinfo{booktitle}{European
  Conference on Computer Vision}, \bibinfo{organization}{Springer}. pp.
  \bibinfo{pages}{392--408}.

\end{thebibliography}



\end{document}